\documentclass[10pt,journal,compsoc]{IEEEtran}

\ifCLASSINFOpdf
\else
\usepackage[dvips]{graphicx}
\fi
\usepackage{url}
\usepackage{booktabs}
\usepackage[colorlinks,linkcolor=blue,anchorcolor=blue,
citecolor=blue]{hyperref}

\usepackage{graphicx}
\usepackage{amsmath,amssymb}
\usepackage{booktabs,tabularx,multirow,threeparttable}
\usepackage{float}
\usepackage{framed}
\usepackage{amssymb}
\usepackage{threeparttable}
\usepackage{multirow}
\usepackage{ragged2e}
\usepackage{enumitem}
\usepackage{makecell}
\usepackage{diagbox}
\usepackage{subfigure}
\usepackage{pifont}
\usepackage[table,xcdraw]{xcolor}

\ifCLASSOPTIONcompsoc
  \usepackage[nocompress]{cite}
\else
  \usepackage{cite}
\fi
\def\ie{\emph{i.e.}}
\def\eg{\emph{e.g.}}

\def\etal{{\em et al.~}}
\usepackage{xcolor}

\newcommand{\figref}[1]{Fig.~\ref{#1}}

\begin{document}

\title{STAR: A First-Ever Dataset and A Large-Scale Benchmark for Scene Graph Generation in Large-Size Satellite Imagery}
\author{
Yansheng Li,~\IEEEmembership{Senior Member,~IEEE}, Linlin Wang, Tingzhu Wang, Xue Yang, Junwei Luo, \\Qi Wang,~\IEEEmembership{Senior Member,~IEEE},Youming Deng, Wenbin Wang, Xian Sun,~\IEEEmembership{Senior Member,~IEEE}, Haifeng Li,~\IEEEmembership{Member,~IEEE}, Bo Dang, Yongjun Zhang,~\IEEEmembership{Member,~IEEE}, Yi Yu, and Junchi Yan,~\IEEEmembership{Senior Member,~IEEE}
\thanks{
\hspace*{\parindent}Yansheng Li, Linlin Wang, Tingzhu Wang, Junwei Luo, Bo Dang and Yongjun Zhang are with School of Remote Sensing and Information Engineering, Wuhan University, Wuhan 430079, China (e-mail: yansheng.li@whu.edu.cn; wangll@whu.edu.cn; tingzhu.wang@whu.edu.cn; luojunwei@whu.edu.cn; bodang@whu.edu.cn; zhangyj@whu.edu.cn). \\
\hspace*{\parindent}Xue Yang is with OpenGVLab, Shanghai AI Laboratory, Shanghai 200030, China (e-mail: yangxue@pjlab.org.cn). \\
\hspace*{\parindent}Qi Wang is with School of Artificial Intelligence, Optics and Electronics (iOPEN), Northwestern Polytechnical University, Xi’an 710072, China (e-mail: crabwq@gmail.com). \\
\hspace*{\parindent}Youming Deng is with Department of Computer Science, Cornell University, Ithaca 14853, United States (e-mail: ymdeng@cs.cornell.edu). \\
\hspace*{\parindent}Wenbin Wang is with College of Computer and Information Technology, Yichang 443002, China Three Gorges University, China (e-mail: wangwenbin@ctgu.edu.cn). \\
\hspace*{\parindent}Xian Sun is with Aerospace Information Research Institute, Chinese Academy of Sciences, Beijing 100190, China (e-mail: sunxian@mail.ie.ac.cn). \\
\hspace*{\parindent}Haifeng Li is with School of Geosciences and Info-Physics, Central South University, Changsha 410083, China (e-mail: lihaifeng@csu.edu.cn). \\
\hspace*{\parindent} Yi Yu is with School of Automation, Southeast University, Nanjing 210096, China (e-mail: yuyi@seu.edu.cn)\\
\hspace*{\parindent}Junchi Yan is with School of Artificial Intelligence, Shanghai Jiao Tong University, Shanghai 200240, China (e-mail: yanjunchi@sjtu.edu.cn).}
}

\markboth{}%
{Shell \MakeLowercase{\textit{et al.}}: Bare Demo of IEEEtran.cls for IEEE Journals}

\IEEEtitleabstractindextext{
	\justify
	\begin{abstract}
Scene graph generation (SGG) in satellite imagery (SAI) benefits promoting understanding of geospatial scenarios from perception to cognition. In SAI, objects exhibit great variations in scales and aspect ratios, and there exist rich relationships between objects (even between spatially disjoint objects), which makes it attractive to holistically conduct SGG in large-size very-high-resolution (VHR) SAI. However, there lack such SGG datasets. Due to the complexity of large-size SAI, mining triplets $<$subject, relationship, object$>$ heavily relies on long-range contextual reasoning. Consequently, SGG models designed for small-size natural imagery are not directly applicable to large-size SAI. This paper constructs a large-scale dataset for SGG in large-size VHR SAI with image sizes ranging from 512 × 768 to 27,860 × 31,096 pixels, named STAR (Scene graph generaTion in lArge-size satellite imageRy), encompassing over 210K objects and over 400K triplets. To realize SGG in large-size SAI, we propose a context-aware cascade cognition (CAC) framework to understand SAI regarding object detection (OBD), pair pruning and relationship prediction for SGG. We also release a SAI-oriented SGG toolkit with about 30 OBD and 10 SGG methods which need further adaptation by our devised modules on our challenging STAR dataset. The dataset and toolkit are available at: \url{https://linlin-dev.github.io/project/STAR}.
 \end{abstract}
\begin{IEEEkeywords}
	Scene graph generation benchmark, large-size satellite imagery, object detection, relationship prediction.
\end{IEEEkeywords}
}
\maketitle

\section{Introduction}\label{sec:introduction}
\begin{figure}[!t]
	\begin{center}
		\includegraphics[width=1.0\linewidth]{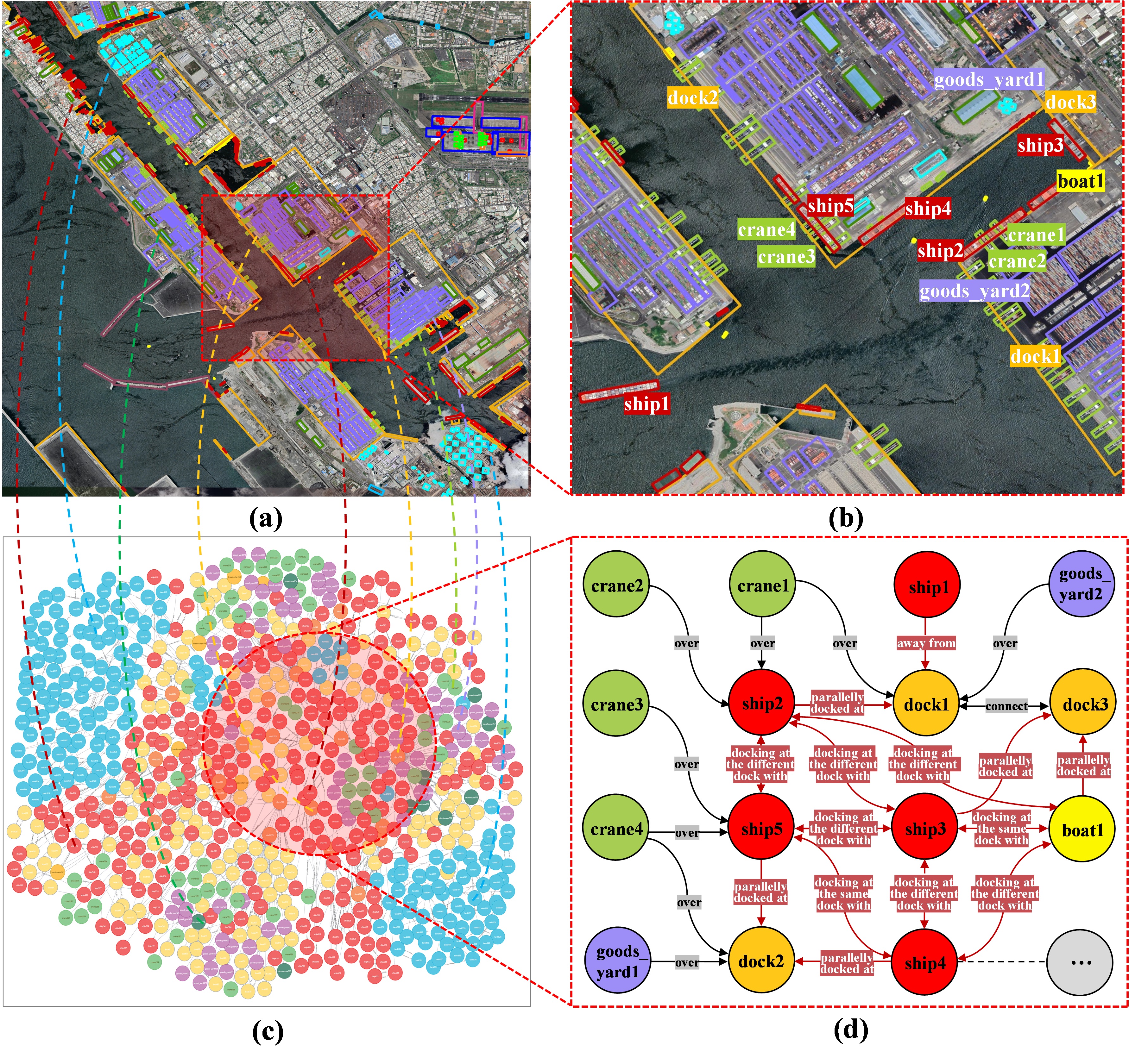}
	\end{center}
    \vspace{-15pt}
	\caption{Illustration of scene graph generation (SGG) in large-size VHR satellite imagery (SAI). (a) and (c) show OBD and SGG results in large-size VHR SAI, respectively. In (d), black arrows denote semantic relationships whose prediction only depends on isolated pairs, but red arrows denote semantic relationships that as inferred from contexts.}
	\label{fig:llustration}
\end{figure}
\IEEEPARstart{S}{cene} graph generation (SGG)~\cite{chang2021comprehensive, cong2023reltr, Sun2023Unbiased} in satellite imagery (SAI) aims to detect objects and predict relationships between objects, where scene graph is a high-level structured representation of SAI via extensive triplets $<$subject, relationship, object$>$. Compared with traditional perceptual understanding tasks in SAI, such as object detection~\cite{xia2018dota,sun2022fair1m, yang2022scrdet++}, scene classification~\cite{cheng2017remote, nogueira2017towards,zhu2021spatial} and semantic segmentation~\cite{hossain2019segmentation, wurm2019semantic, li2021learning}, SGG in SAI works towards cognitively mining the high-value knowledge contained in SAI (\eg, in \figref{fig:llustration}, $<$ship1, away from, dock1$>$ can be inferred by the spatial layout between ship1 and dock1 and the wake trailing of ship1). SGG in SAI benefits cognitive interpretation tasks in SAI, like image retrieval~\cite{johnson2015image}, image captioning~\cite{yang2019auto} and image visual question answering~\cite{Li_2019_ICCV}, and further serves various applications, \eg traffic planning~\cite{reinartz2006traffic, smigaj2023monitoring}, circuit layout~\cite{matikainen2016remote}. 

As illustrated in \figref{fig:llustration}(a) and \figref{fig:llustration}(b), large-size VHR SAI allows for the retention of detailed information on small objects without compromising the integrity on large objects, which provides mandatory support for predicting rich relationships between multi-scale objects. However, to the best of our knowledge, there is a significant lack of both datasets and approaches for SGG in large-size VHR SAI. Therefore, there is an urgent need to explore datasets and approaches around SGG in large-size VHR SAI to pursue the cognitive understanding of SAI.

Table \ref{table:available STAR} shows existing datasets for the SGG task in SAI~\cite{chen2021message,li2021semantic}, which face limitations concerning the number of objects, number of triplets, and image size. Moreover, they tend to provide horizontal bounding box (HBB) annotations with too much background disturbance, hurting the recognition of relationships between man-made objects with distinct oriented boundaries in SAI. For the object detection (OBD) task in SAI, while the datasets for OBD in SAI~\cite{li2020object,ding2021object,sun2022fair1m} are continually improving in object annotation granularity as well as annotation volume, they lack relationship annotations. Additionally, they necessitate considerable improvements in terms of image size, object diversity, and scene complexity, all of which are crucial for the SGG task in SAI. Given these constraints, it appears unrealistic to train a robust SGG model using these limited datasets. Specifically, some specific characteristics should be considered while annotating the triplets $<$subject, relationship, object$>$ for the SGG task in SAI: \textbf{(i) Large-size VHR SAI spanning diverse scenarios}. In SAI, objects exhibit large variations, and there exist rich relationships between objects in various complex scenarios, even between those not spatially intersecting. Large-size VHR SAI retains the detailed information of small objects while simultaneously preserving the integrity of large objects, which makes the predictions of relationships between objects at different scales/ranges feasible. \textbf{(ii) Fine-grained objects with precise annotations}. Given the top-down perspective inherent in SAI, objects such as ships, trucks and runways often appear in arbitrary orientations. Therefore, precise object annotations are necessary for accurate localization. Notably, fine-grained object categories are also indispensable for practical applications of SGG task in SAI. For instance, when investigating parking violations $<$car, incorrectly parked on, truck\underline{ }parking$>$, it is crucial to distinguish not only between fine-grained vehicle categories (truck/car) for subject, but also between fine-grained parking categories (truck\underline{ }parking/car\underline{ }parking) for object. \textbf{(iii) High-value relationships considering contextual reasoning}. Unlike natural imagery, the relationships between objects in SAI are rotationally invariant (\ie, they do not change with the rotation of the image), and a significant number of object pairs containing high-value relationships are spatially disjoint in SAI. Furthermore, many meaningful relationships between objects are heavily context-dependent, and the same types of object pairs in SAI may correspond to different categories under different contexts.

\begin{table*}[t!]
\small
\centering
\caption{Comparison of the STAR dataset with other datasets for OBD and SGG tasks in SAI (only for datasets sourced from satellite platforms and covering GSD $<$ 1m). The comparison includes image size, object annotation type, ground sampling distance (GSD), object information (the number of instances, the number of classes, and the number of objects per image (OPI)), relationship information (the number of triplets, the number of classes, and the number of relationships per image (RPI)), source, and public.}
\renewcommand{\arraystretch}{1.3} 
\vspace{-10pt}
\resizebox{1.0\textwidth}{!}{
\begin{tabular}{ccccccccccccc}
\hline
\multirow{3}{*}{\textbf{Task type}} & \multirow{3}{*}{\textbf{Dataset}} & \multirow{3}{*}{\textbf{Image size}} & \multirow{3}{*}{\begin{tabular}[c]{@{}c@{}}\textbf{Annotation}\\ \textbf{type}\end{tabular}} & \multirow{3}{*}{\textbf{GSD}} & \multicolumn{3}{c}{\textbf{Object}} & \multicolumn{3}{c}{\textbf{Relationship}} & \multirow{3}{*}{\textbf{Venue}} & \multirow{3}{*}{\textbf{Public}} \\ \cline{6-11}
 &  &  &  &  & \begin{tabular}[c]{@{}c@{}}Num. of\\ instances\end{tabular} & \begin{tabular}[c]{@{}c@{}}Num. of\\ classes\end{tabular} & \begin{tabular}[c]{@{}c@{}}Num.of\\ OPI\end{tabular} & \begin{tabular}[c]{@{}c@{}}Num. of\\ triplets\end{tabular} & \begin{tabular}[c]{@{}c@{}}Num.of\\ classes\end{tabular} & \begin{tabular}[c]{@{}c@{}}Num.of\\ RPI\end{tabular} &  &  \\ \hline
\multirow{7}{*}{OBD} & NWPU VHR-10~\cite{cheng2016learning} & $\sim$1,000 & HBB & 0.08$\sim$2 & 3,775 & 10 & 4.8 & - & - & - & TGRS'16 & \textbf{\checkmark} \\
 & DOTA-v1.0~\cite{xia2018dota} & 800$\sim$13,000 & OBB & 0.5 & 188,282 & 15 & 67.1 & - & - & - & CVPR'18 & \textbf{\checkmark} \\
 & HRRSD~\cite{zhang2019hierarchical} & 152$\sim$10,569 & HBB & 0.15$\sim$1.2 & 55,740 & 13 & 2.6 & - & - & - & TGRS'19 & \textbf{\checkmark} \\
 & DIOR~\cite{li2020object,cheng2022anchor} & 800 & OBB & 0.5$\sim$30 & 190,288 & 20 & 8.1 & - & - & - & ISPRS'20 & \textbf{\checkmark} \\
 & DOTA-v2.0~\cite{ding2021object} & 800$\sim$20,000 & OBB & 0.5 & 1,973,658 & 18 & 159.2 & - & - & - & TPAMI'21 & \textbf{\checkmark} \\
 & FAIR1M~\cite{sun2022fair1m} & 1,000$\sim$10,000 & OBB & 0.3$\sim$0.8 & 1,020,000 & 37 & 66.8 & - & - & - & ISPRS'22 & \textbf{\checkmark} \\
 & GLH-Bridge~\cite{li2024learning} & 2,048$\sim$16,384 & OBB & 0.3$\sim$1 & 59,737 & 1 & 10.0 & - & - & - & TPAMI'24 & \textbf{\checkmark} \\ \hline
\multirow{3}{*}{SGG} & RSSGD~\cite{li2021semantic} & 224 & HBB & - & - & 39 & - & - & 13 & - & ISPRS'21 & × \\
 & GRTRD~\cite{chen2021message} & 600 & HBB & 0.5 & 19,904 & 12 & 6.2 & 18,602 & 33 & 5.8 & GRSL'21 & × \\
 & \textbf{STAR (Ours)} & \textbf{512 $\sim$31,096} & \textbf{OBB} & \textbf{0.15$\sim$1.0} & \textbf{219,120} & \textbf{48} & \textbf{172.1} & \textbf{400,795} & \textbf{58} & \textbf{314.8} & - & \textbf{\checkmark} \\ \hline
\end{tabular}}
\label{table:available STAR}
\end{table*}

To address the dataset scarcity, we introduce a first-ever dataset for \textbf{S}cene graph genera\textbf{T}ion in l\textbf{A}rge-size satellite image\textbf{R}y \textbf{(STAR)}, which contains more than 210,000 objects and more than 400,000 triplets for SGG in large-size VHR SAI. In this dataset, \textbf{(i)} SAI with a spatial resolution of 0.15m to 1m is collected, covering 11 categories of complex geospatial scenarios associated closely with human activities worldwide (\eg, airports, ports, nuclear power stations and dams). \textbf{(ii)} By the guidance of human experts, all objects are classified into 48 fine-grained categories and annotated with oriented bounding boxes (OBB), and all relationships are annotated in accordance with 8 major categories including 58 fine-grained categories. \textbf{(iii)} All object pairs and their contained relationships are one-to-many annotated, and all relationship annotations are absolute (unaffected by imagery rotation). In conclusion, STAR has significant advantages over existing OBD datasets and SGG datasets in SAI. 

Specifically, this paper opens up a new challenging and practical topic: \textbf{SGG in large-size VHR SAI}. Considering the characteristics of large-size VHR SAI, potential solutions should overcome three main problems: \textbf{(i) OBD}. Multi-class OBD in large-size VHR SAI presents significant challenges, primarily due to the holistic detection for multi-class objects with diverse object scales (\eg, same category: bridges, docks; different categories: "airplanes and aprons", "ships and docks"), extreme aspect ratios (\eg, bridges, runways and gravity\underline{ }dams), and arbitrary orientations (\eg, trucks, ships, airplanes, bridges, runways). These factors make it challenging to holistically detect these objects while preserving the complete details necessary for accurate relationship prediction. \textbf{(ii) Pair pruning}. Large-size VHR SAI usually contains a large number of objects and a plethora of object pairs (the number of which grows exponentially as the number of objects), which inevitably results in excessive computation for exhaustive methods based on all object pairs. Thus, it is necessary to prune object pairs without losing high-value pairs. Notably, the triplets are usually partially annotated in reality (\ie, many non-annotated object pairs still carry meaningful relationships), which suggests that some approaches treating pair sparsity as a binary classification problem (\ie, annotated pairs are treated as positive samples and non-annotated pairs are treated as negative samples) are not reasonable. \textbf{(iii) Relationship prediction}. Different from small-size natural imagery, there are a large number of object pairs containing rich knowledge, which are even spatially disjoint in large-size SAI. A few relationships in large-size SAI can be directly predicted by isolated pairs, while most high-value relationships need to be inferred with the help of context. As illustrated in \figref{fig:llustration}(c) and \figref{fig:llustration}(d), only simple triplets (\eg, $<$crane1, over, dock1$>$) in large-size SAI can be predicted when relying only on isolated pairs. However, more complex triplets can be inferred when the context of the pairs is introduced (\eg, $<$ship2, parallelly docked at, dock1$>$ and $<$ship5, parallelly docked at, dock2$>$ can be inferred as $<$ship2, docking at the different dock with, ship5$>$).

In this paper, we propose a context-aware cascade cognition (CAC) framework to understand SAI at three levels: OBD, pair pruning and relationship prediction. \textbf{First}, a holistic multi-class object detection network (HOD-Net) incorporating multi-scale context is adopted to detect multi-scale objects in large-size VHR SAI. \textbf{Then}, to reduce the complexity of predicting pairs without losing high-value pairs, a pair proposal generation (PPG) network via adversarial reconstruction is designed to obtain contextual interaction pairs containing rich knowledge. \textbf{Finally}, considering the dependence of relationship prediction in large-size SAI on the contexts of object pairs, a relationship prediction network with context-aware messaging (RPCM) is proposed to comprehensively predict the relationship types.

In summary, the STAR dataset and CAC framework proposed in this paper aim to establish a large-scale benchmark for SGG in large-size VHR SAI. Under this benchmark, new algorithms can be further developed. The  contributions are:
		\begin{itemize}[leftmargin=*]
			\item 
   			We propose STAR, a first-ever dataset for SGG in large-size VHR SAI. Containing more than 210K objects and over 400K triplets across 1,273 complex scenarios globally, it provides more challenging and practical testbed for the community.

			\item
			To address SGG in large-size VHR SAI, a context-aware cascade cognition (CAC) framework is established where the HOD-Net is devised for holistic detection of multi-class multi-scale objects in large-size SAI, and then the pair proposal generation (PPG)  network and relationship prediction network with context-aware messaging (RPCM) network are designed to retain high-value pairs as well as to predict relationship types of these pairs.
			
            \item
            We release an open-source toolkit, encompassing about 30 OBD methods and 10 SGG methods, designed to accommodate OBD and SGG in large-size VHR SAI. The versatile toolkit unifies the SGG process in natural imagery and SAI, offering flexibility and ease of use. Our toolkit will provide valuable contributions to the development of the SGG community.  

		\item
	    Rigorous experiments on the STAR dataset, employing state-of-the-art algorithms, underscore the indispensability of this dataset and the efficacy of the algorithms. The experimental results showcased in this paper serve as a large-scale benchmark for future research in this direction.
		\end{itemize}

\begin{figure*}[!tb]
	\begin{center}
		\includegraphics[width=1.00\linewidth]{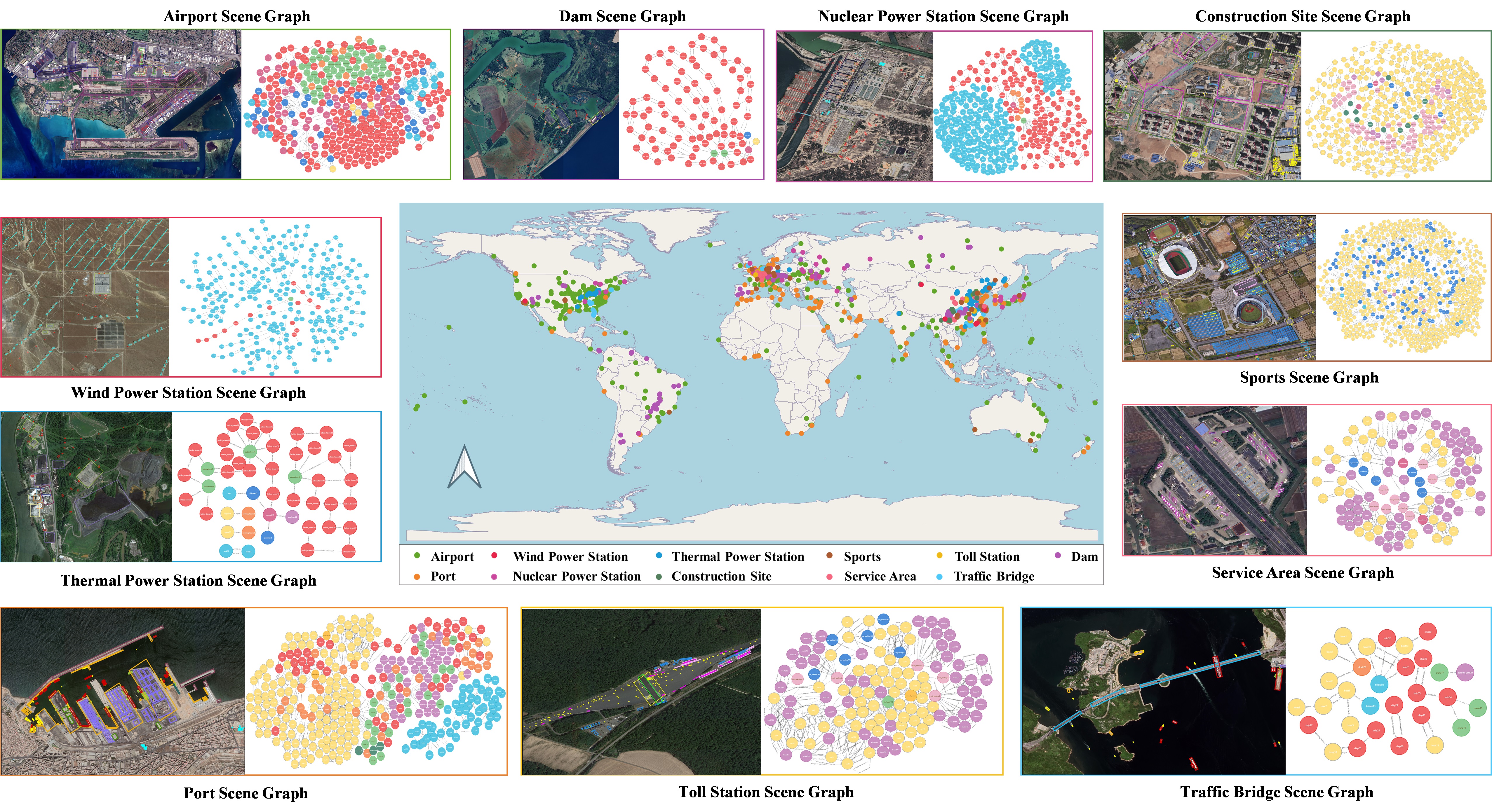}
	\end{center}
	\vspace{-10pt}
	\caption{The geographical distribution map of the sampled images from the proposed STAR dataset.}
	\label{fig:distribution}
\end{figure*}

\section{Related Work}\label{sec:related}

\subsection{Scene Graph Generation Datasets}
Scene graph has attracted extensive attention from many researchers as a powerful tool for the understanding and reasoning analysis of images~\cite{schuster2015generating, Kim_2019_CVPR, Li_2019_ICCV, yang2019auto}. Various datasets have been proposed for SGG in recent years. 
\subsubsection{Scene Graph Generation Datasets in Natural Images}\label{subsec:SGGdataNI}

Visual Phrases~\cite{sadeghi2011recognition} is an early dataset for visual relationship detection, which is a task similar to SGG, and it has 8 object categories and 13 relationship categories. RW-SGD~\cite{johnson2015image} may be the first dataset for SGG, is constructed by gathering 5,000 images from YFCC100m~\cite{thomee2015new} and Microsoft COCO datasets~\cite{lin2014microsoft}, containing 93,832 object instances and 112,707 relationship instances. VRD~\cite{lu2016visual} is constructed for the task of visual relationship detection, which has 100 object categories from 5,000 images and contains 37,993 relationship instances. Visual Genome (VG)~\cite{krishna2017visual} is a large-scale scene graph dataset, which consists of 108,077 images with tens of thousands of object and relationship categories. Later, VG variants~\cite{xu2017scene,liang2019vrr} are gradually introduced by researchers to address its key shortcomings. Specifically, VG-150~\cite{xu2017scene} retains only the most common 150 object categories and 50 predicate categories to reflect a more realistic scenario. VrR-VG~\cite{liang2019vrr} argues that many predicates in VG-150 could be easily estimated by statistics, and then re-filters the original VG categories. However, there are still cases of redundancy and ambiguity of predicates in VrR-VG. Similar shortcomings are also found in the popular datasets GQA~\cite{hudson2019gqa}, Open Images V4 and V6~\cite{kuznetsova2020open}. In addition, the co-occurrence of multiple relationships between subject-object pairs is common in the real world, but most previous datasets treat edge prediction as single-label categorization.

\subsubsection{Scene Graph Generation Datasets in SAI}\label{subsec:SGGdataSAI}

Compared to the natural imagery field, the development of SGG in SAI field has been relatively slow. SGG datasets in SAI~\cite{chen2021message,li2021semantic,lin2022srsg,li2024aug} with detailed annotation and considering the characteristics of SAI are rare. GRTRD~\cite{chen2021message} may be the first SGG dataset in SAI, which focuses on 12 kinds of objects and contains 19,904 object instances and 18,602 relationship instances. RSSGD~\cite{li2021semantic} is a SAI-oriented SGG dataset constructed based on the remote sensing caption dataset RSICD~\cite{lu2017exploring}, which has about 39 object categories and 16 relationship categories. Unfortunately, the image size of GRTRD dataset is 600×600 pixels, and the image size of RSSGD dataset is only 224×224 pixels. Since the large-size VHR SAI scenarios are cropped into small image blocks, the objects and relationships contained in each image are very sparse, and many object pairs containing high-value relationships are corrupted, which makes it difficult to support high-level cognitive understanding of SAI. In addition, the objects in the above dataset are annotated by HBB with too much background disturbance, which does not provide accurate information for recognizing relationships between regular man-made objects with distinct oriented boundaries in large-size VHR SAI. To meet the need for cognitive understanding of SAI, a large-scale SGG dataset named STAR for large-size VHR SAI is constructed, which contains multiple complex scenarios and meaningful relationships. In Table \ref{table:available STAR}, we give the statistics of popular OBD datasets and SGG datasets in SAI. Compared to small-scale datasets for the SGG task in SAI, such as RSSGD dataset and GRTRD dataset, our STAR outperforms them by at least an order of magnitude in both the number of objects and relationships on all/single images. For OBD datasets in SAI~\cite{cheng2016learning,zhang2019hierarchical,ding2021object} lacking relationship annotations, our STAR dataset provides effective improvements in both object category diversity and scene complexity. The STAR dataset, a large-scale and comprehensive SGG dataset, supports both the OBD task and the SGG task based on the HBB-based and OBB-based detectors in large-size VHR SAI, serving as an invaluable resource for cognitive understanding of SAI.

\subsection{Scene Graph Generation Methods}
\begin{figure*}[!tb]
	\begin{center}
		\includegraphics[width=1.00\linewidth]{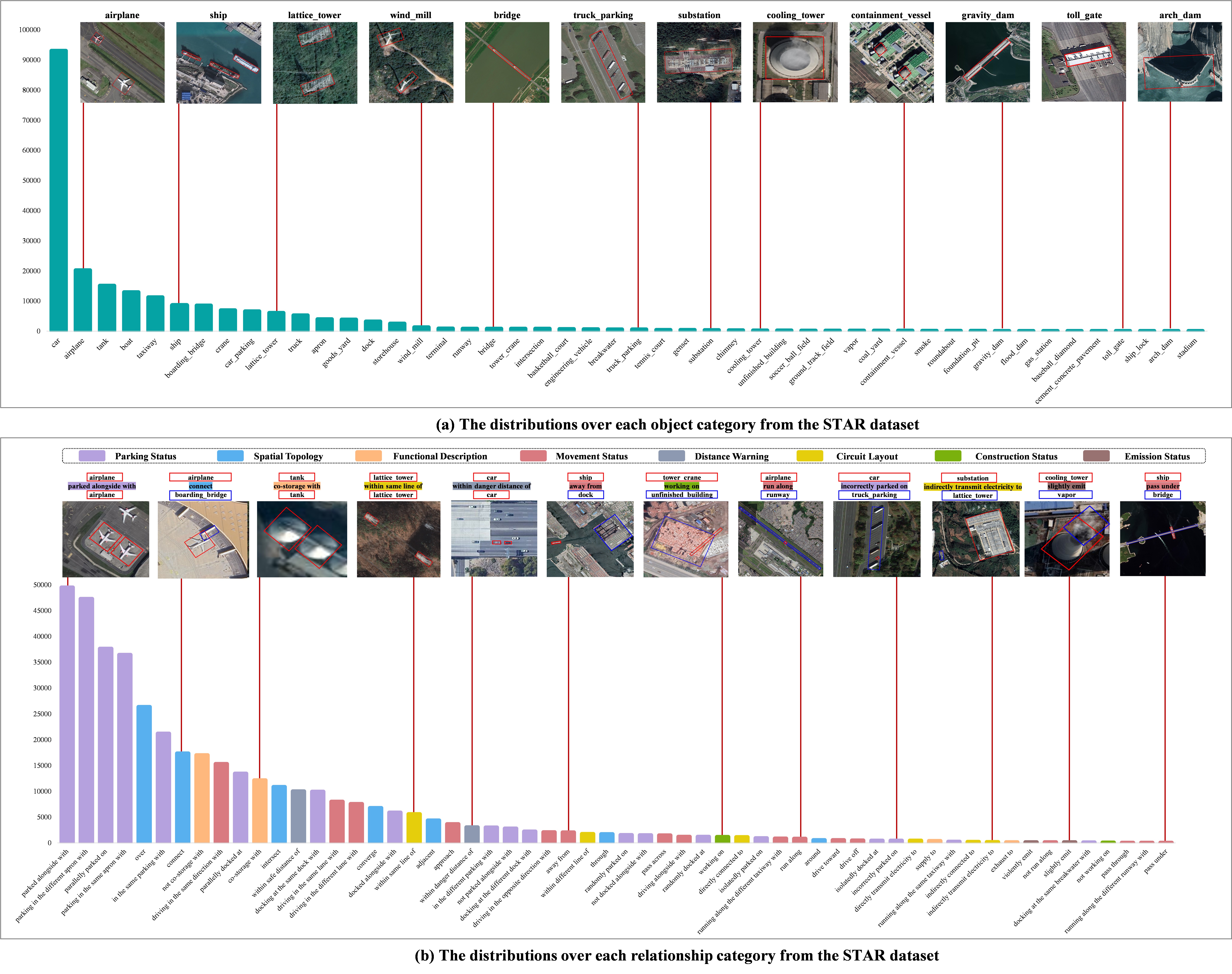}
	\end{center}
	\vspace{-15pt}
	\caption{Statistics and visualization of objects (a) and relationships (b) from the STAR dataset. The relationships are color-coded to show parking statu, spatial topology and functional description, movement status, distance warning, circuit layout, construction status and emission status. Some typical objects and interesting triplets are visualized.}
	\label{fig:object+relationships}
 \vspace{-8pt}
\end{figure*}
\subsubsection{Scene Graph Generation Methods in Natural Imagery}\label{subsec:SGGNI}
Existing SGG methods for natural imagery have been dominated by the two-stage process consisting of OBD and relationship prediction between objects. Given the need for compatibility between OBD models and relationship prediction models within the SGG framework, the majority of existing SGG models, in alignment with the literature~\cite{tang2020unbiased,dong2022stacked}, adopt Faster R-CNN as the standard for the OBD task. To generate high-quality scene graphs from images, a series of works have been explored in several directions, such as task-specific~\cite{xu2017scene,zellers2018neural,tang2019learning,zheng2023prototype,yoon2023unbiased} and dataset-specific~\cite{tang2020unbiased,chen2022resistance,deng2022hierarchical}. Notably, aggregating contextual information between objects has been shown to be effective for SGG. Xu \etal\cite{xu2017scene} first combined contextual information to refine the features of objects and relationships by using GRU to pass messages between graphs. Based on the discovery that there are some inherent relationship patterns in subgraphs, Motifs~\cite{zellers2018neural} constructed a new global context computing mechanism. A series of context-based studies~\cite{yang2018graph,lin2020gps,li2021bipartite} have consecutively explored the messaging mechanisms of objects and relationships. Most recently, HetSGG~\cite{yoon2023unbiased} attempted to capture the relationships between objects in more detail by proposing a heterogeneous graph and proved its effectiveness. However, the above methods of exhaustive enumeration based on all pairs tend to result in excessive computation, which makes it difficult to run under common computational resources (\eg, 24 GB of memory in a single GPU) for large-size VHR SAI with a large number of objects and more object pairs.

\begin{figure}[!tb]
	\begin{center}
            \includegraphics[width=0.98\linewidth]{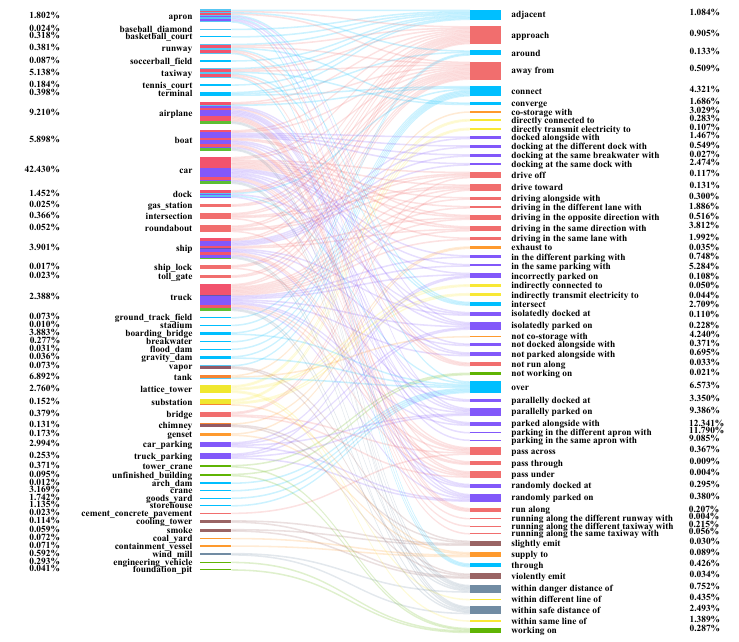}
	\end{center}
	\vspace{-15pt}
	\caption{Interaction mapping between objects and relationships. There are eight colors in the figure, which represent the types of eight relationships: parking status (in purple), spatial topology (in blue) and functional description (in orange), movement status (in red), distance warning (in grey), circuit layout (in gold), construction status (in green) and emission status (in brown). The values on either side indicate the proportion of each object category and relationship category, respectively.}
	\label{fig:interaction}
\end{figure}

\subsubsection{Scene Graph Generation Methods in SAI}\label{subsec:SGGSAI}

Limited by a few semantic understanding datasets in SAI, there are only a few approaches related to SGG in SAI. Shi and Zou~\cite{shi2017can} proposed a framework for SAI captioning by exploring whether machines can automatically generate human-like linguistic descriptions of SAI, and this work provides an initial exploration of high-level understanding of SAI. Chen \etal\cite{chen2021message} proposed a geospatial relationship triplet representation dataset (GRTRD) based on the characteristics of high-resolution SAI and then adopted the "object-relationship" message-passing mechanism to realize the prediction of geographic objects and geospatial relationships. Li \etal\cite{li2021semantic} proposed a multi-scale semantic fusion network (MSFN) for SGG in SAI based on their dataset RSSGD, which integrates global information through a graph convolution network to improve the accuracy of SGG. However, the above methods are mainly designed for small-size SAI with HBB and do not take into account the characteristics of large-size VHR SAI, such as the extreme aspect ratio of the objects, the great scale variation among the objects, and the random orientation of the objects. Furthermore, due to the limitation of the dataset, these methods do not have the ability to predict relationships between objects separated at long-range in large-size VHR SAI.


\section{Details of the STAR Dataset}\label{sec:Details}

\subsection{Image Collection and Annotation}
\subsubsection{Image Collection}\label{subsec:Collection}


To meet the needs of practical applications, images in the STAR dataset are collected from Google Earth, with a spatial resolution ranging from 0.15m to 1m. Considering the value of cognitive understanding of large-size VHR SAI scenarios in practical applications closely related to human activities, such as transportation, energy, and life, we collected scenarios from more than 1,200 airports, ports, wind power stations, nuclear power stations, thermal power stations, construction sites, sports, service areas, toll stations, traffic bridges, and dams from all over the world. Unlike objects in natural imagery, objects in SAI are oriented in a variety of directions. Therefore, HBB-based annotation cannot provide accurate spatial information for oriented objects. To wrap objects more accurately and develop more suitable algorithms for oriented objects in SAI, all object instances in the STAR dataset are annotated with OBB. Samples of annotated instances in the STAR dataset can be seen in \figref{fig:distribution}.

\subsubsection{Annotation Criteria}\label{subsec:Criteria}
The construction of the SGG dataset for large-size VHR SAI mainly contains two tasks: object annotation and relationship annotation. Referring to the category system of mainstream OBD datasets in SAI, and combining with the actual requirements of the SGG task and some downstream tasks such as visual question answer and image caption, we carry out detailed interpretation and value analysis of the object categories appearing in the above scenarios, and finally choose 48 objects categories as shown in \figref{fig:object+relationships}(a). Specifically, we select to annotate "smoke" and "vapor" because they can be used as a basis for judging the working status of power plants. In addition, the collection and annotation of objects like "containment\underline{ }vessel", "lattice\underline{ }tower", "substation" and "gravity\underline{ }dam" are expected to play an important role in SAI-based electricity research~\cite{matikainen2016remote}. 

For the relationship description, unlike the ``human-eye view" perspective of natural imagery, the ``birds-eye view" perspective in SAI makes the semantic relationships between objects absolute, so we discard the orientation words such as ``east", ``south", ``west", and ``north". Moreover, we have learned a lesson from the ambiguous labeling of SGG datasets in natural imagery and removed relationship descriptions whose semantics are obviously ambiguous. After careful screening, the semantic relationships are defined into 8 major categories: ``distance warning", ``spatial topology", ``functional description", ``circuit layout", ``movement status", ``emission status", ``construction status" and ``parking status", which contain 58 subcategories as in \figref{fig:object+relationships}(b).

\figref{fig:interaction} gives the interaction mapping between objects and relationships. There are eight colors, representing the eight major categories of relationships consistent with \figref{fig:object+relationships}(b). It can be seen that almost every relationship category is associated with multiple object categories, which reflects the complex interactions between objects in large-size VHR SAI.

\subsubsection{Annotation Management}\label{management}
STAR dataset has a standardized dataset construction pipeline: pre-annotation and rules refinement stage, large-scale detailed annotation stage, and quality-checking stage. In the pre-annotation stage, 6 experts in the SAI field are first invited to formulate annotation rules, and then the main authors pre-annotate the images with 11 types of scenarios, which are examined and evaluated by these experts, thereby forming a detailed object and relationship annotation document. During the large-scale annotation stage, the main authors provide comprehensive training to 9 professionals with rich interpretation experience in SAI, and each image is annotated by 2 professionals and checked by 1 professional at the same time. During the quality-checking stage, cross-sampling and revisions were performed by main authors.

Specifically, we use RoLabelImg4 to manually generate fine-grained oriented labeled boxes for the objects. Each object is represented by four clockwise-aligned angular coordinates $\left (x_{1},y_{1},x_{2},y_{2},x_{3},y_{3},x_{4},y_{4} \right)$ and category labels, and each triplet is represented by $<$subject, relationship, object$>$. For a more meaningful annotation, all pre-annotated pairs (both long-range and short-range) are selected based on the interaction graph in \figref{fig:interaction} during the relationship annotation process.

\subsection{Data Characteristics and Analysis}\label{subsec:CandA}
\subsubsection{Data Characteristics}\label{subsec:Characteristics}
Compared with the existing OBD dataset in SAI, the STAR dataset has unique properties on task diversity, category diversity, and scale diversity. Sparse high-value pairs, high intra-class variation and inter-class similarity are also important characteristics of the STAR dataset.

\begin{figure}[!tb]
	\begin{center}
		\includegraphics[width=1.0\linewidth]{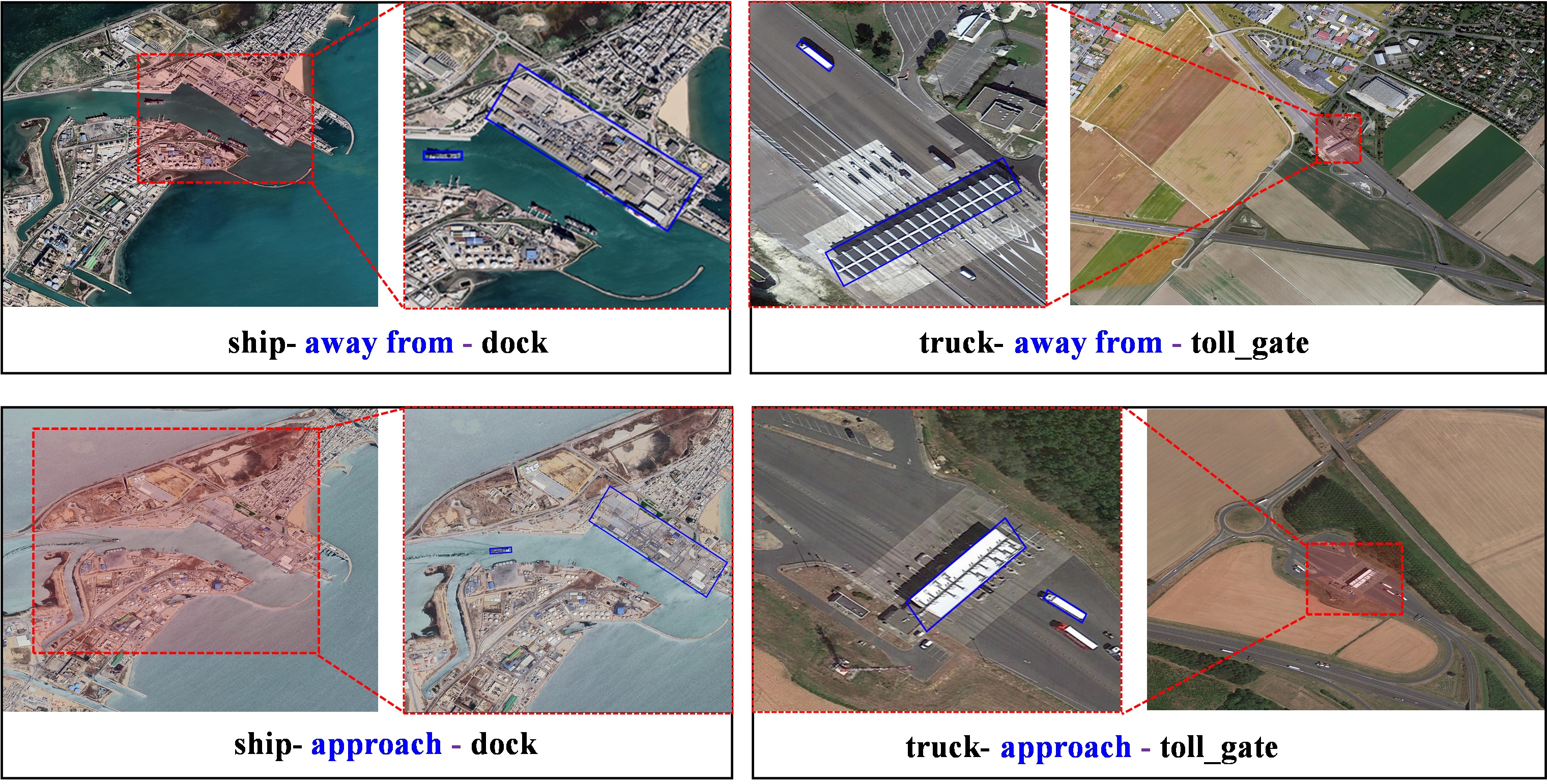}
	\end{center}
 \vspace{-15pt}
	\caption{Examples of intra-class variations and inter-class similarities in relationship on the STAR dataset.}
	\label{fig:Examples}
\end{figure}

\begin{figure*}[!tb]
	\begin{center}
		\includegraphics[width=0.98\linewidth]{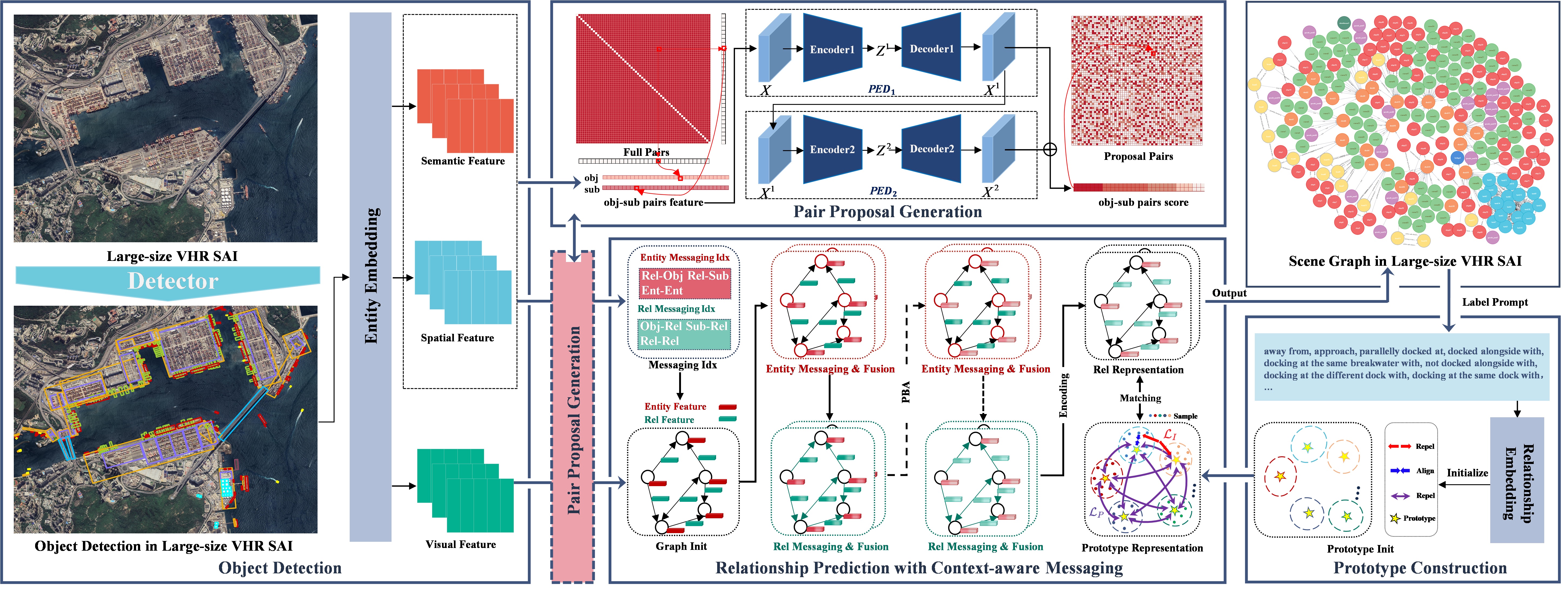}
	\end{center}
	\vspace{-15pt}
	\caption{The overall architecture of the CAC framework. First, multi-scale objects in large-size VHR SAI are holistically detected by HOD-Net. Then, contextual interaction pairs containing rich knowledge are selected by the PPG network via adversarial reconstruction. Finally, high-value relationships between subject and object are predicted by the RPCM network.}
	\label{fig:pipeline1}
\end{figure*}

\textbf{Task diversity.} STAR supports both the OBD and SGG tasks in large-size VHR SAI. After processing the annotated images, it can further satisfy the various task requirements like OBD based on HBB/OBB in large-size VHR SAI, SGG based on HBB/OBB in large-size VHR SAI.

\textbf{Category diversity.} STAR has various scene categories, various object categories, and various relationship categories. Specifically, it contains 11 categories of VHR SAI scenarios over the world, 48 categories of important objects, and 58 categories of high-value relationships, which will bring new opportunities for cognitive understanding of SAI.

\textbf{Scale diversity.} In the STAR dataset, there are many typical objects with extreme aspect ratios (\eg, runways, bridges, gravity\underline{ }dams), and there are large scale variations between objects of the same category (\eg, airplanes, ships) and different categories (\eg, ``boats and docks", ``cars and car\underline{ }parkings") that contain rich knowledge, which will bring great challenges to the OBD and SGG tasks.

\textbf{Sparse high-value pairs.} There are an enormous number of objects in a large-size SAI. If relationship prediction is performed for all object pairs, more than $N(N-1)$ triplets will appear for $N$ objects in the case of one-to-many relationships (\eg, in the STAR dataset where the number of objects in one image is up to 6,800, the possible triplets will be more than 40,000,000), which will cause memory overflow under common computational resources. In fact, not all object pairs are of interest to us, and only a few of them contain high-value relationships. Therefore, it will be a new challenge to select high-value pairs from the numerous pairs for the SGG task in large-size VHR SAI.

\textbf{High intra-class variation and inter-class similarity.} Intra-class variation of relationships is mainly due to the differential appearance and diverse subject-object pairs. The inter-class similarity of relationships arises from similar appearance representations, but manifests itself differently for different relationship categories. As shown in \figref{fig:Examples}, for the object pairs of the same type, the relationship labels between them are non-unique (\eg, $<$ship, away from, dock$>$, $<$ship, approach, dock$>$), and for the object pairs of different types, their relationship labels may be the same (\eg, $<$ship, away from, dock$>$, $<$truck, away from, toll\underline{ }gate$>$).

\subsubsection{Data Statistics and Visualization}\label{subsec:Statistics}
The statistical results of object and relationship annotations from the STAR dataset are shown in \figref{fig:object+relationships}(a) and \figref{fig:object+relationships}(b), respectively. It can be seen that there is an obvious long-tail effect for both object and relationship categories in the STAR dataset, which is also consistent with reality. The highest-frequency objects are common small objects such as ``cars", ``airplanes", ``tanks" and ``boats". The highest-frequency relationships are mainly related to the parking states, such as "parked alongside with", ``parking in the different apron with", ``parallelly parked on" and ``parking in the same apron with".

\section{The Proposed Method}\label{sec:method2}
\figref{fig:pipeline1} presents the overall architecture of the proposed CAC framework, which mainly contains three parts: OBD, pair pruning and relationship prediction. Considering the drastic changes in the scales and aspect ratios of objects in large-size VHR SAI, a HOD-Net that flexibly integrates multi-scale contexts is proposed for the holistic detection of multi-class objects. Facing the problem of memory overflow caused by pair redundancy in large-size VHR SAI under common computational resources, we propose the PPG network via adversarial reconstruction to obtain object pair ranking scores, which can efficiently sift out contextual interaction pairs containing rich knowledge under the condition of no-negative samples. Considering the dependence of relationship prediction in large-size SAI on the contexts of object pairs, the RPCM network for predicting relationship types of these pairs is proposed, which enhances the cognitive ability of the model by fusing the contexts of objects and relationships in triplets. Based on the three levels in the CAC framework, the definition of the SGG task in large-size VHR SAI is presented here. Specifically, viewing the object entities as nodes of the scene graph, and the relationships between the entities as directed edges connecting different entities, the SGG task in large-size VHR SAI can be described as:
\begin{equation}
\text{\small $P(G_{SA} \mid I) = P(E \mid I) P(E_{S}, E_{O} \mid E, I) P(R \mid E_{S}, E_{O}, I),$}
\label{eq:G}
\end{equation}
where $I$ denotes large-size SAI, $G_{SA}$ denotes the scene graph in large-size SAI, $E$ is object entities, and $R$ is the relationships between subjects $E_{S}$ and objects $E_{O}$.

\subsection{Oriented Object Detection in Large-size VHR SAI}\label{subsec:OOBD}

OBD in large-size SAI is an indispensable basis for SGG in large-size SAI, which directly affects the prediction accuracy of the triplets. Compared with small-size or low-resolution imagery, large-size VHR SAI can not only accommodate more complete large objects (\eg, runways, docks), but also greatly retain important details of small objects (such as the head and tail of a car, the head and tail of an airplane), which can help to mine more high-value knowledge in SAI. In processing large-size VHR SAI, mainstream deep learning-based OBD methods, whether supervised~\cite{yang2018automatic, ding2019learning, yang2019scrdet, hou2023g} or weakly supervised~\cite{yang2023h2rbox,yu2023h2rboxv2,luo2024pointobb,yu2024point2rbox}, adopt a cropping strategy~\cite{xia2018dota,akyon2022slicing}, which results in large objects, such as runways, being cut off. In addition to the cropping strategy, some OBD methods tackle the original large-size imagery with a fixed-window down-sampling strategy~\cite{chen2023coupled}, resulting in the loss of massive detail information of the image. Recently, HBD-Net for single-class bridge detection in large-size VHR SAI is proposed~\cite{li2024learning}, and the SDFF architecture in HBD-Net effectively solves the large scale variations in single-class bridge detection by fusing multi-scale contexts via the dynamic image pyramid (DIP) of the large-size image. However, this single-class OBD only considers the intra-class variations of the objects, which is not applicable in the multi-class OBD task with both intra-class and inter-class variations. 

 To address the aforementioned problem, it extends HBD-Net to the holistic multi-class OBD in large-size VHR SAI, named HOD-Net whose total loss of the HBB-based and OBB-based detectors in large-size VHR SAI is defined as:
\begin{align}
 \mathcal{L}_{O}&=\sum_{m=1}^{\mathcal{M}}\left(\frac{1}{\Gamma^{m}}\!\sum_{i\in \Delta^{m}}\!\mathcal{L}_{i}^{cls}\!+\!\frac{1}{{\Gamma^{m}}^{+}}\sum_{j\in {\Delta^{m}}^{+}}\!w^{reg}_{j}\mathcal{L}_{j}^{reg}\right), \\
\mathcal{L}_{H}&=\sum_{m=1}^{\mathcal{M}}\left(\frac{1}{\Gamma^{m}}\!\sum_{i\in \Delta^{m}}\!\mathcal{L}_{i}^{cls}\!+\!\frac{1}{{\Gamma^{m}}^{+}}\!\sum_{j\in {\Delta^{m}}^{+}}\!\mathcal{L}_{j}^{reg}\right), 
\end{align}
\noindent
where $m$ means the layer of the DIP, $m\in [1,\mathcal{M}]$. $\Delta^{m}$ and ${\Delta^{m}}^{+}$ are the set of all samples and the set of positive samples, respectively. $\Gamma^{m}$ and ${\Gamma^{m}}^{+}$ indicate the total number of all samples and positive samples in the $m$-th layer, respectively. ${w^{reg}_{j}}$ denotes the regression weight obtained using the SSRW strategy~\cite{li2024learning}. Regression loss $\mathcal{L}_{j}^{reg}$ is set with reference to the smoothed L1 loss defined in~\cite{girshick2015fast}.

To enable the model to adaptively focus on different object categories in different layer, a hierarchical adaptive weighted strategy is designed in classification loss $\mathcal{L}^{cls}_{i} = -t_{c}log(e^{\mathbf{\Psi}_{c}^{m}}/\sum_{c=1}^{\mathcal{C} }e^{\mathbf{\Psi}_{c}^{m}})$. Specifically, $t_{c}$ represents the one-hot vector, and $\mathbf{\Psi}_{wei}^{m} = \mathbf{w}^{m}\odot \mathbf{\Phi}^{m}$ denotes the class-weighted score. $\mathbf{w}^{m}=[w_{1}^{m},...,w_{\mathcal{C}}^{m}]$ is the learnable weight, and $\mathbf{\Phi}^{m}=[\phi_{1}^{m},...,\phi_{\mathcal{C}}^{m}]$ denotes  the confidence of the network. $\mathcal{C}$ is the number of object categories, and $\odot$ is the element-wise product.

\subsection{Pair Proposal Generation via Adversarial Reconstruction}\label{subsec:PPGvia}
Given $N$ objects in the image, the number of subject-object pairs is $N(N-1)$, which exhibits exponential growth as $N$ increases. There are a huge amount of subject-object pairs in a large-size VHR SAI, and the pair redundancy problem will cause the model to be ineffective under common computational resources due to memory overflow. Some researchers regard pair pruning as a binary classification problem for the SGG task in natural imagery, \ie, annotated object pairs are regarded as positive samples, and non-annotated object pairs are regarded as negative samples. It is worth noting that it is difficult to exhaust the annotation of triplets for large-size VHR SAI, and most of the non-annotated object pairs still have high-value semantic relationships, so it is unreasonable to directly regard the non-annotated object pairs as negative samples. To address this problem, this paper proposes a pair proposal generation (PPG) network via adversarial reconstruction, which can realize the ranking of triplets according to their score based on their containment of high-value knowledge without negative samples, and effectively addresses the issue of selecting meaningful pairs with contextual associations in the case of only positive samples for the SGG task in large-size VHR SAI.


To enable the model to autonomously learn the reconstruction ability for high-value pairs without negative samples, the PPG network is divided into two-stage adversarial training with pair encoder-decoder ($PED$), because the auto-encoder can achieve stability and reconstruct the inputs well during adversarial training. The PPG network contains two encoder networks: encoder1 ($E_{1}$), and encoder2 ($E_{2}$) as well as two decoder networks: decoder1 ($D_{1}$), and decoder ($D_{2}$), which constitute the two auto-encoders $PED_{1}$ and $PED_{2}$, as shown in \figref{fig:pipeline1}. Two $PED$ are trained to reconstruct the input feature $X$,  which represents the union of spatial and semantic features for object pairs.

Similar to the gaming strategy in generative adversarial network (GAN)~\cite{goodfellow2014generative}, $PED_{1}$ and $PED_{2}$ are trained in an adversarial manner, in which $PED_{1}$ tries to deceive $PED_{2}$, while $PED_{2}$ tries to distinguish whether the data is the input $X$ or $X^{1}$ reconstructed by $PED_{1}$. During training stage, the feature $X$ is squeezed into the latent space $Z$ by $E_{1}$, and then reconstructed by $D_{1}$, which can be represented as $D_{1}(E_{1}(X))$. $PED_{2}$ is trained by $D_{2}(E_{2}(X^{1}))$  to distinguish whether the data is the original input $X$ or from $PED_{1}$, during which $PED_{1}$ maintains learning to deceive $PED_{2}$. Overall,  $PED_{1}$ and $PED_{2}$ similar to minimal-maximal games:
\begin{align}
\zeta = \min_{PED_{1} } \max_{PED_{2} }G\left \| X-PED_{2}(PED_{1}(X)) \right \|_{2},
\label{eq:p12}
\end{align}

The training objective of $PED_{1}$ is to minimize the reconstruction error $\left \| X_{i} -X_{i}^{1}  \right \| _{2}$ to learn the latent representation of the data, and the training objective of $PED_{2}$ is to reduce the reconstruction error $\left \| X_{i} -X_{i}^{2}  \right \| _{2}$ to a minimum. As the iterations $n$ are constantly changing, the loss is:
\begin{align}
\mathcal{L}_{PED}=\frac{1}{n} \left \| X -X^{1}  \right \|_{2}  +(1-\frac{1}{n})\left \| X -X^{2}  \right \|_{2},
\label{eq:ped}
\end{align}

In the inference stage, $n$ in Eq. (5) is set to 2, and then the top $k_{1}$ proposal pairs are selected as input indices of subject-object pairs for relationship prediction.

\subsection{Relation Prediction with Context-Aware Messaging}\label{subsec:rpwithcm}

For large-size VHR SAI, any two objects can often be directly associated through significant relationships or indirectly associated with other medium objects, so the introduction of context-aware messaging from objects and relationships is essential to enhance the cognitive ability of the model. In addition, numerous possible subject-object pairs present different visual appearances, resulting in large intra-class variation among the same relationships, which pose critical challenges for SGG in large-size VHR SAI. This paper proposes a relationship prediction network with context-aware messaging (RPCM), which mainly consists of two components: a progressive bi-context augmentation component and a prototype-guided relationship learning component. The former uses context messaging from nodes and edges to augment the local representations of objects and relationships. Meanwhile, the latter continuously optimizes and updates the semantic prototype under instance-level and prototype-level constraints. Ultimately, accurate relationship prediction is achieved by matching the predicted relationship representation with the prototype representation in the same semantic space.

\subsubsection{Progressive Bi-context Augmentation Network}\label{subsec:PBA}

\begin{figure}[!tb]
	\begin{center}
		\includegraphics[width=1.0\linewidth]{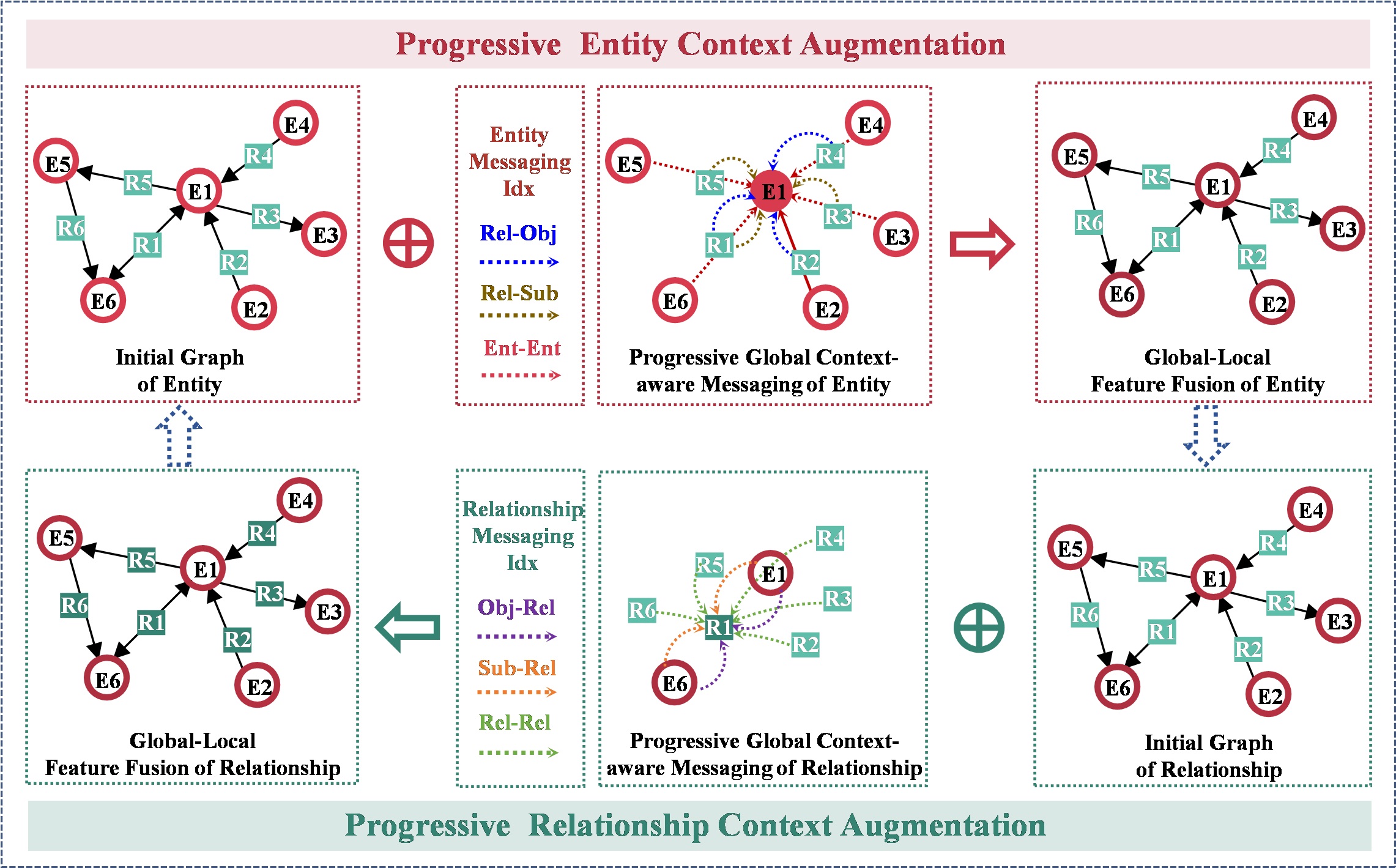}
	\end{center}
 \vspace{-15pt}
	\caption{The proposed PBA module: the details of messaging and fusion for one progressive context augmentation of entity-relationship.}
	\label{fig:PBA}
\end{figure}

In the sparse graph formulated based on the proposed PPG network, the semantic representations of both objects and relationships are influenced by neighboring objects and relationships. However, the interplay between object-object, object-relationship, and relationship-relationship interactions may exhibit substantial diversity in different contexts. Attention graph network~\cite{velickovic2018graph}, which leads the model to focus on reliable context information by assigning the corresponding weights to different regions of an image, has shown its effectiveness for SGG in natural imagery~\cite{lin2020gps,yoon2023unbiased}. Nonetheless, due to the complexity of large-size VHR SAI, conventional SGG models often exhibit deficiencies in relationship context inference capabilities when processing SAI. Moreover, although the integration of global context information assists in enhanced relationship recognition, more fine-grained discriminative features are primarily extracted from the triplets themselves. An overabundance of global context information may induce significant confusion. In our work, a novel progressive bi-context augmentation (PBA) component is proposed. As shown in \figref{fig:PBA}, PBA mainly consists of entity-relationship progressive global context-aware messaging and global-local feature fusion. The progressive global context-aware messaging module for entity-relationship adaptively learns global information from neighboring entities and relationships using a progressive strategy. The global-local feature fusion module fuses the collected global and local features to obtain augmented features. After several iters of nodes and edges in the sparse graph, the dual optimization and updating of entity and relationship representations are accomplished.

 \textbf{Progressive Global Context-aware Messaging.} Considering that it is easy to confuse when introducing too much unnecessary context information, we adopt a progressive training strategy to solve this problem. In this strategy, all entities and relationships first collect reliable information from their direct neighboring nodes and edges via the attention graph network, and then learn their indirectly related contexts with the help of neighboring nodes and edges, so that each entity and relationship can learn more reliable contextual information without introducing extra noise.
 
Sparse scene graphs constructed based on the proposed PPG network can provide multiple types of messaging indexes for entities and relationships. For $i+1$ iters, different types of messaging (\ie, entity-entity, relationship-subject, and relationship-object) are performed on entities to learn context-aware information of entities:
\begin{align}
\widehat{F}_{i+1}^{e}=\sigma (w^{ee}{F}_{i}^{e}\alpha^{ee}+w^{rs}{F}_{i}^{r} \alpha ^{rs} +w^{ro}{F}_{i}^{r}\alpha ^{ro}),
\label{eq:objectness}
\end{align}
where $\sigma(\cdot)$ denotes the sigmoid activation function~\cite{nair2010rectified}. $\alpha^{ee}, \alpha^{rs}$ and $\alpha^{ro}$ represent the importance of messaging from entity to entity, relationship to subject, and relationship to object, respectively. By learning to adjust $\alpha$~\cite{velickovic2018graph}, each iteration leads to a change in attention and affects subsequent iters. $w^{ee}, w^{rs}$ and $w^{ro}$ denote the weight matrix of the directed messaging from entity to entity, relationship to subject, and relationship to object, respectively. ${F}_{i}^{e}$ and ${F}_{i}^{r}$ are entity  and relationship representation, respectively.

Similarly, different types of messaging (\ie, relationship-relationship, subject-relationship, object-relationship) on relationships are used to learn the global contextual information of relationships:
\begin{align}
\widehat{F}_{i+1}^{r}=\sigma (w^{rr}{F}_{i}^{r}\alpha^{rr}+w^{sr}{F}_{i}^{e} \alpha ^{sr} +w^{or}{F}_{i}^{e}\alpha ^{or}),
\label{eq:relationness}
\end{align}



\textbf{Global-Local Feature Fusion.} Introducing reliable global information helps to better recognize relationships, but more fine-grained discriminative features mainly come from local. Therefore, we adopt a global-local feature fusion strategy to capture reliable global information and maintain the local more discriminative features. Specifically, the context augmented feature of entity ${F}_{i+1}^{e}$ can be obtained by fusing the context-aware feature of entity $\widehat{F}_{i+1}^{e}$ from global and the entity feature ${F}_{0}^{e}$ from local. Similarly, the context augmentation feature of relationship ${F}_{i+1}^{r}$ can be obtained. After $L$ iters, the predicted relationship representations $Rel$ with context-aware augmentation are updated. Note that the predicted relationship representations in Sec.~\ref{subsec:PBA} are optimized and updated simultaneously with the prototype representations in Sec.~\ref{subsec:PRL}.

\subsubsection{Prototype-guided Relationship Learning}\label{subsec:PRL} 
Prototype learning allows SGG models to focus on all categories equally in training, and is effective in alleviating the category imbalance problem of datasets~\cite{zheng2023prototype}. However, previous linear classifier-based methods often lack the ability to model data at different semantic levels and suffer from a lag in representational capabilities (\eg, $<$ship, away from, dock$>$ and $<$ship, approach, dock$>$ correspond to different relationship labels, but it is clear that $<$ship, away from, dock$>$ and $<$ship, approach, dock$>$ are closer than $<$truck, away from, gas\underline{ }station$>$ in the visual representation). We propose a prototype-guided relationship learning method that utilizes category labels as prompts to help generate learnable relationship prototypes, and prototype representations updated dynamically by instance-level constraints and prototype-level constraints to help the SGG model enhance the discriminability between prototypes.

\textbf{Prototype Construction.} To construct more discriminative prototype representations, word embeddings (Glove) of different category labels $[label_{1}, label_{2},..., label_{C}]$ are utilized to construct initial class-specific semantic prototypes $\mathbf{T}=[t_{1}, t_{2}, ..., t_{C}]$. After obtaining the initial prototype representations, they are fed into a learnable prototype encoder $Enc_{p}$, during which the prototype representations $Pro =Enc_{p}(\mathbf{T})$ are allowed to be updated and matched with the predicted relationship representations $Rel$ in the same semantic space by $r=MAP(Rel)$ and $p=MAP(Rro)$, where $MAP(\cdot)$ is two-layer MLPs. 

\textbf{Instance-wise Matching Loss.} To enable the model to learn more fine-grained category variation, the matching constraints between samples and prototypes are injected into the learning objective during instance-level matching to achieve positive samples closer to the prototypes and negative samples further away from the prototypes. Specifically, we obtain instance-level loss $\mathcal{L}_{IC}$ and $\mathcal{L}_{ID}$ based on cosine distance and euclidean distance, respectively:
\begin{align}
\mathcal{L}_{IC}&=-\log\frac{\exp(< \overline{r},\overline{p}> /\tau)}{ \textstyle \sum_{i=0}^{C}\exp(<\overline{r},\overline{p}_{i}> /\tau) }, \\
\mathcal{L}_{ID}&=\max(0,q^{+}-q^{-}+\gamma _{1}),
\end{align}
where $q^{-}=\left \| r^{Neg}-\overline{p}_{i}\right \|_{2}^{2}, q^{+}=\left \| r^{true}-\overline{p}_{i}\right \|_{2}^{2}$. $r^{true}$ denotes positive samples with label annotations, and $r^{Neg}$ is the negative samples sampled in order of distance with the exclusion of positive samples.

\textbf{Prototypical Matching Loss.} The purpose of prototype contrast learning is to make samples of the same category tightly clustered around the corresponding prototype to form tight clusters, with clusters corresponding to different prototypes pulling away from each other. To achieve this goal, we utilize the semantic structure captured by relationship prototypes to realize prototype contrast learning loss $\mathcal{L}_{PC}$ and loss $\mathcal{L}_{PD}$.
\begin{align}
\mathcal{L}_{PC}&=\left \| \overline{p}\cdot\overline{p}^{T}\right \| _{2,1}, \\
\mathcal{L}_{PD}&=max(0,-\mathcal{N}_{select}^{k}(\left \| \overline{p_{i}}-\overline{p_{j} } \right \|_{2}^{2})+\gamma _{2}),
\label{eq:ploss}
\end{align}
where $\mathcal{N}_{select}^{k}(D_{M})$ denotes the selection of the top $k$ smallest values in the distance matrix $D_{M}$. 

The overall training loss $\mathcal{L}$ of our RPCM is as follows, without any bias on the weight of each term:
\begin{equation}
 \mathcal{L} = \mathcal{L} _{IC}+\mathcal{L} _{ID}+ \mathcal{L} _{PC}+ \mathcal{L} _{PD},
\label{eq:overall}
\end{equation}

In the inference stage, the prototype type with the highest cosine similarity $s_{i}$ is used for relationship prediction:
\begin{equation}
R_{class} =\underset{i}{\text{argmax}} ({s_{i}\mid s_{i}=< \overline{r},\overline{p_{i} } > /\tau}),
\label{eq:o}
\end{equation}
where $\overline{r}$ and $\overline{p}$ are both regularization operations.

\begin{table*}[tb!]
\renewcommand{\arraystretch}{1.5} 
	\centering
	\caption{Baseline results (\%) of HBB-based and OBB-based detectors on STAR test set. b\_b denotes boarding\_bridge, l\_t denotes lattice\_tower, s\_l is ship\_lock, and g\_d represents gravity\_dam. All experiments are based on the standard `1x' (12 epochs) training schedule. $^\dagger$ means that Swin-L~\cite{liu2021swin} is used and others indicates ResNet50~\cite{he2016deep} as the backbone. \underline{Underline} indicates the base detector for subsequent lines.}
\vspace{-8pt}
	\resizebox{1.0\textwidth}{!}{
    \centering
		\begin{tabular}{c|clc|cccccccccccccccc}
\hline
\multirow{2}{*}{\textbf{OBD}}  & \multicolumn{2}{c}{\multirow{2}{*}{\textbf{Detectors}}} & \multirow{2}{*}{\textbf{Venue}} & \multirow{2}{*}{\textbf{ship}} & \multirow{2}{*}{\textbf{boat}} & \multirow{2}{*}{\textbf{truck}} & \multirow{2}{*}{\textbf{car}} & \multirow{2}{*}{\textbf{airplane}} & \multirow{2}{*}{\textbf{crane}} & \multirow{2}{*}{\textbf{b\_b}} & \multirow{2}{*}{\textbf{tank}} & \multirow{2}{*}{\textbf{l\_t}} & \multirow{2}{*}{\textbf{bridge}} & \multirow{2}{*}{\textbf{runway}} & \multirow{2}{*}{\textbf{s\_l}} & \multirow{2}{*}{\textbf{dock}} & \multirow{2}{*}{\textbf{g\_d}} & \multirow{2}{*}{...} & \multirow{2}{*}{\textbf{mAP}} \\
                      & \multicolumn{2}{c}{}  &  &    &   &    &    &   &  &   &    &   &     &    &   &   &   &   &   \\ \hline
\multirow{8}{*}{HBB}  & \multicolumn{2}{c}{{Faster R-CNN}~\cite{ren2015faster}}                          & NeurIPS’15 & 30.7  & 7.2 & 26.2 & 38.1 & 54.3  & 23.6 & 24.7 & 21.1 & 21.9 & 5.3 & 0.0  & 0.0  & 4.2  & 0.0  & ... & 32.2  \\
                      & \multicolumn{2}{c}{RetinaNet~\cite{lin2017focal}}                          & ICCV'17  & 26.1  & 4.6 & 25.6 & 35.5 & 52.6  & 20.1 & 21.0 & 18.4 & 21.9  & 3.3 & 0.0 & 0.0   & 2.8 & 0.0    & ...   & 24.6   \\
                      & \multicolumn{2}{c}{{Cascade R-CNN}~\cite{cai2019cascade}}                         & TPAMI'19 & 32.0 & 7.8 & 29.3 & {\textbf{39.1}} & {\textbf{55.0}} & 26.4  & 25.6 & 20.9  & 22.1  & 6.0  & 0.0  & 0.0 & 5.6  & 0.0   &  ... & 32.4  \\
                      & \multicolumn{2}{c}{{FCOS}~\cite{tian2020fcos}}                          & TPAMI'20  & 20.0  & 2.3 & 23.7  & 34.6 & 48.0  & 14.8  & 17.4  & 17.6  & 11.5 & 1.2 & 0.0  & 0.0 & 2.0  & 0.0 & ... & 20.8  \\
                      & \multicolumn{2}{c}{{TOOD}~\cite{feng2021tood}}                          & ICCV'21  & 29.3  & 5.5 & 30.8   & 38.4  & 53.6  & 23.5 & 24.8 & 19.8  & 20.6   & 4.0  & 0.0  & 3.3  & 3.8  & 1.2   & ...  & 30.1  \\
                      & \multicolumn{2}{c}{GCL~\cite{chen2023coupled}}                          & KBS'23   & 30.6 & 7.2 & 26.6 & 38.1  & 54.3 & 23.5  & 24.5   & 21.1 & 21.8  & 5.4  & 6.6  & 0.0 & 5.0  & 2.2 & ...  & 34.4  \\
                      & \multicolumn{2}{c}{HOD-Net (Ours)}              & -                       & 32.5 & 7.6   & 28.5 & 37.5 & 53.3 & 25.5 & 24.5  & {\textbf{21.8}} & 22.8  & 9.7 & 9.3  & 0.9  & 12.3 & 37.7  & ...   & 45.2 \\
                      & \multicolumn{2}{c}{HOD-Net$^\dagger$ (Ours)}              & -                       & {\textbf{35.4}} & {\textbf{11.8}} & {\textbf{31.9}}   & 36.7 & 54.0 & {\textbf{33.5}}  & {\textbf{27.1}}   & 20.8 & {\textbf{23.1}}  & {\textbf{15.7}}   & {\textbf{20.9}}  & {\textbf{4.7}} &{\textbf{ 15.4}} & {\textbf{40.3}}   & ...   & {\textbf{53.7}} \\ \hline
\multirow{28}{*}{OBB}  & \multicolumn{2}{c}{\underline{Deformable DETR}~\cite{zhu2020deformable}} & ICLR'21 & 18.1 & 5.5 & 15.8 & 42.5 & 85.1 & 12.6 & 32.7 & 51.1 & 21.2 & 0.5 & 0.0 & 0.0 & 2.1 & 0.0 & ... & 17.1 \\
& \multicolumn{2}{c}{ARS-DETR~\cite{zeng2024ars}} & TGRS'24 & 44.6 & 16.4 & 36.1 & 60.8 & 88.6 & 40.5 & 59.0 & 54.7 & 41.6 & 8.2 & 0.0 & 0.0 & 7.8 & 0.0 & ... & 28.1 \\
& \multicolumn{2}{c}{\underline{RetinaNet}~\cite{lin2017focal}}                  & ICCV'17                  & 39.3                  & 14.3                  & 36.7                   & 65.3                 & 88.5                      & 42.9                   & 43.1                  & 50.8                  & 44.8                  & 2.1                     & 0.0                     & 0.0                   & 5.2                   & 0.0                   & ...                  & 21.8                 \\
                      & \multicolumn{2}{c}{ATSS~\cite{zhang2020bridging}} & CVPR'20                       & 38.4 & 15.5 & 31.4 & 67.6 & 89.0 & 35.6 & 39.2 & 52.6 & 43.2 & 5.1 & 0.0 & 0.0 & 5.1 & 0.0 & ... & 20.4 \\
                      & \multicolumn{2}{c}{KLD~\cite{yang2021learning}}                        & NeuIPS'21                  & 43.1                  & 12.7                  & 39.1                   & 65.7                 & 88.7                      & 47.2                   & 50.0                  & 51.2                  & 48.0                  & 3.9                     & 0.0                     & 0.0                   & 12.7                  & 0.0                   & ...                  & 25.0                 \\
                      & \multicolumn{2}{c}{GWD~\cite{yang2021rethinking}}                        & ICML'23                  & 42.5                  & 14.5                  & 38.4                   & 65.8                 & 89.1                      & 52.2                   & 54.0                  & 51.0                  & 49.2                  & 4.4                     & 0.0                     & 0.0                   & 12.6                  & 0.0                   & ...                  & 25.3                 \\
                      & \multicolumn{2}{c}{KFIoU~\cite{yang2023kfiou}}                      & ICLR'23                  & 41.6                  & 14.1                  & 39.5                   & 67.1                 & 89.4                      & 47.7                   & 54.6                  & 51.4                  & 47.0                  & 8.3                     & 0.0                     & 0.0                   & 9.2                   & 4.5                   & ...                  & 25.5                 \\
                      & \multicolumn{2}{c}{DCFL~\cite{xu2023dynamic}} & CVPR'23 & 47.3 & 10.2 & 45.6 & 69.7 & 89.2 & 56.8 & 52.9 & 55.2 & 49.2 & 13.2 & 0.0 & 0.0 & 11.3 & 2.3 & ... & 29.0 \\
                      & \multicolumn{2}{c}{R$^3$Det~\cite{yang2021r3det}} & AAAI'21 & 44.0 & 15.3 & 45.3 & 68.6 & 89.6 & 52.5 & 43.5 & 53.6 & 47.5 & 4.3 & 0.0 & 0.0 & 11.2 & 0.0 & ... & 23.7 \\
                      & \multicolumn{2}{c}{S$^2$A-Net~\cite{han2021align}} & TGRS'21 & 46.6 & 13.9 & 45.1 & 69.3 & 89.7 & 52.2 & 52.4 & 57.9 & 48.9 & 13.5 & 0.0 & 0.0 & 11.7 & 0.0 & ... & 27.3 \\
                      & \multicolumn{2}{c}{\underline{RepPoints}~\cite{yang2019reppoints}} & ICCV'19 & 26.3 & 5.6 & 27.1 & 62.5 & 89.0 & 40.3 & 35.3 & 54.2 & 38.6 & 9.1 & 0.0 & 0.0 & 4.9 & 0.0 & ... & 19.7 \\
                      & \multicolumn{2}{c}{CFA~\cite{guo2021beyond}} &  CVPR'21 & 45.6 & 15.1 & 44.3 & 66.8 & 88.8 & 57.3 & 48.6 & 56.1 & 46.4 & 2.9 & 0.0 & 0.0 & 13.7 & 0.0 & ... & 25.1 \\
                      & \multicolumn{2}{c}{Oriented RepPoints~\cite{li2022oriented}} & CVPR'22 & 48.0 & 16.9 & 43.8 & 67.9 & 89.3 & 59.0 & 49.7 & 55.1 & 46.5 & 9.1 & 0.0 & 0.0 & 11.4 & 0.0 & ... & 27.0 \\
                      & \multicolumn{2}{c}{G-Rep~\cite{hou2023g}} & RS'23 & 43.3 & 17.2 & 47.2 & 68.5 & 89.2 & 53.3 & 54.1 & 54.7 & 40.5 & 12.5 & 0.0 & 0.0 & 4.6 & 0.0 & ... & 26.9 \\
                      & \multicolumn{2}{c}{SASM~\cite{hou2022shape}} & AAAI'22 & 40.6 & 10.4 & 42.6 & 68.1 & 89.0 & 56.3 & 41.8 & 51.0 & 44.1 & 14.5 & 0.0 & 0.0 & 14.3 & 0.0 & ... & 28.2 \\
                      & \multicolumn{2}{c}{\underline{FCOS}~\cite{tian2019fcos}}                       & ICCV'19                  & 43.3                  & 12.7                  & 44.9                   & 68.2                 & 89.5                      & 56.9                   & 53.6                  & 52.9                  & 44.4                  & 12.7                    & 0.0                     & 0.0                   & 12.9                  & 0.0                   & ...                  & 28.1                 \\
                      & \multicolumn{2}{c}{CSL~\cite{yang2020arbitrary}} & ECCV'20 & 43.8 & 12.6 & 45.5 & 67.3 & 89.1 & 55.5 & 52.1 & 49.9 & 44.3 & 10.3 & 0.0 & 0.0 & 8.9 & 0.0 & ... & 27.4 \\
                      & \multicolumn{2}{c}{PSC~\cite{yu2024boundary}} & TPAMI'24 & 44.6 & 18.0 & 44.3 & 66.2 & 89.6 & 56.4 & 55.4 & 53.1 & 44.8 & 16.1 & 0.0 & 0.0 & 13.9 & 0.0 & ... & 30.5 \\
                      & \multicolumn{2}{c}{H2RBox-v2~\cite{yu2023h2rboxv2}} & NeurIPS'23 & 43.7 & 12.2 & 44.0 & 67.7 & 89.3 & 50.7 & 51.4 & 49.2 & 45.4 & 15.7 & 0.0 & 0.0 & 12.2 & 0.0 & ... & 27.3 \\
                      & \multicolumn{2}{c}{\underline{Faster R-CNN}~\cite{ren2015faster}}               & NeurIPS’15                  & 50.4                  & 20.9                  & 48.1                   & 68.0                 & 89.6                      & 20.9                   & 54.4                  & 55.0                  & 48.6                  & 18.8                    & 0.0                     & 0.0                   & 14.0                  & 0.0                   & ...                  & 32.6                 \\
                      & \multicolumn{2}{c}{Gliding Vertex~\cite{xu2020gliding}}             & TPAMI'20                 & 50.6                  & 23.8                  & 38.6                   & 68.1                 & 89.4                      & 56.9                   & 54.7                  & 54.2                  & 45.9                  & 17.8                    & 0.0                     & 0.0                   & 14.9                  & 0.0                   & ...                  & 30.7                 \\
                      & \multicolumn{2}{c}{RoI Transformer~\cite{ding2019learning}}            & CVPR'19                  & 56.6                  & 26.2                  & 50.3                   & 69.1                 & 89.9                      & 59.7                   & 63.4                  & 57.3                  & 51.0                  & 15.8                    & 0.0                     & 0.0                   & 13.6                  & 0.0                   & ...                  & 35.7                 \\
                      & \multicolumn{2}{c}{ReDet~\cite{han2021redet}}                      & CVPR'21                  & 60.1                  & 25.5                  & 50.0                   & 69.9                & {\textbf{90.1}}                      & 68.0                   & {\textbf{63.7}}                  & {\textbf{60.1}}                  & 54.4                  & 25.2                    & 0.0                     & 0.0                   & 18.6                  & 0.0                   & ...                  & 39.1                 \\
                      & \multicolumn{2}{c}{Oriented R-CNN~\cite{xie2021oriented}}             & ICCV'21                  & 57.6                  & 26.6                  & 51.6                   & 67.8                 & 89.5                      & 65.0                   & 62.8                  & 54.1                  & 49.1                  & 21.2                    & 0.0                     & 0.0                   & 14.8                  & 0.1                   & ...                  & 33.2                 \\
                      & \multicolumn{2}{c}{GCL~\cite{chen2023coupled}}             & KBS'23                  & 57.1                  & 26.6                  & 51.5                   & 67.8                 & 89.4                      & 64.2                   & 62.2                  & 54.1                  & 49.1                  & 21.1                    & 0.0                     & 0.0                   & 15.1                  & 0.1                   & ...                  & 33.7                 \\
                      & \multicolumn{2}{c}{LSKNet-S~\cite{li2023large}}   &  ICCV'23 & 59.0 & 25.5 & 53.2 & \textbf{70.9} & 89.7 & 65.5 & 62.4 & 53.6 & 55.0 & 24.7 & 0.0 & 0.0 & 19.5 & 11.4 & ... & 37.8 \\
                      & \multicolumn{2}{c}{PKINet-S~\cite{cai2024poly}}   &  CVPR'24 & 55.2 & 22.4 & 48.0 & 67.5 & 89.9 & 59.1 & 56.3 & 50.7 & 47.4 & 21.1 & 0.0 & 0.0 & 19.0 & 0.2 & ... & 32.8 \\

                      & \multicolumn{2}{c}{HOD-Net (Ours)}              & -                       & 57.2                    & 23.5                    & 49.4                     & 64.9                   & 89.5                        & 65.5                     & 56.6                    & 56.5                    & 56.2                    & 25.9                      & 11.3                     & 3.7                   & 20.7                    & 48.6                   & ...                  & 43.6                   \\
                      & \multicolumn{2}{c}{HOD-Net$^\dagger$ (Ours)}              & -                       & {\textbf{65.2}}                  & {\textbf{32.6}}                  & {\textbf{53.4}}                   & 66.4                 & 89.7                      & {\textbf{69.3}}                   & 63.1                  & 59.3                  & {\textbf{57.6}}                  & {\textbf{38.7}}                    & {\textbf{35.7}}                    & {\textbf{46.9}}                  & {\textbf{29.4}}                  & {\textbf{78.6}}                  & ...                  & {\textbf{55.9}}                 \\ \hline
\end{tabular}}
	\label{table:object30}
\end{table*}

\begin{table*}[tb!]
\renewcommand{\arraystretch}{1.5} 
	\centering
	\caption{Results (\%) of different training strategies on STAR test set. b\_b denotes boarding\_bridge, l\_t denotes lattice\_tower, s\_l is ship\_lock, and g\_d represents gravity\_dam. All experiments are based on the standard `1x' (12 epochs) training schedule. Swin-L~\cite{liu2021swin} is adopted as the backbone.}
 \vspace{-8pt}
	\resizebox{1.0\textwidth}{!}{
    \centering
		\begin{tabular}{c|cl|cccccccccccccclc}
			\hline
 & \multicolumn{2}{c|}{} &  &  &  &  &  &  &  &  &  &  &  &  &  &  & \multicolumn{1}{c}{} &  \\
\multirow{-2}{*}{\textbf{OBD}} & \multicolumn{2}{c|}{\multirow{-2}{*}{\textbf{Detectors}}} & \multirow{-2}{*}{\textbf{ship}} & \multirow{-2}{*}{\textbf{boat}} & \multirow{-2}{*}{\textbf{truck}} & \multirow{-2}{*}{\textbf{car}} & \multirow{-2}{*}{\textbf{airplane}} & \multirow{-2}{*}{\textbf{crane}} & \multirow{-2}{*}{\textbf{b\_b}} & \multirow{-2}{*}{\textbf{tank}} & \multirow{-2}{*}{\textbf{l\_t}} & \multirow{-2}{*}{\textbf{bridge}} & \multirow{-2}{*}{\textbf{runway}} & \multirow{-2}{*}{\textbf{s\_l}} & \multirow{-2}{*}{\textbf{dock}} & \multirow{-2}{*}{\textbf{g\_d}} & \multicolumn{1}{c}{\multirow{-2}{*}{...}} & \multirow{-2}{*}{\textbf{mAP}} \\ \hline
 & \multicolumn{2}{c|}{Faster R-CNN$^\dagger$~\cite{ren2015faster} (Resizing)} & 7.5 & 0.0 & 12.6 & 1.3 & 1.8 & 3.4 & 0.3 & 0.6 & 0.4 & 6.9 & \textbf{29.5} & 4.7 & 10.1 & \textbf{41.6} & ... & 33.7 \\
  & \multicolumn{2}{c|}{Faster R-CNN$^\dagger$~\cite{ren2015faster} (Cropping)} & \textbf{35.6} & 9.5 & 29.8 & \textbf{37.1} & \textbf{53.1} & 31.6 & \textbf{27.2} & \textbf{20.9} & 22.3 & 14.0 & 0.0 & 0.0 & 10.8 & 10.2 & ... & 42.6 \\
\multirow{-3}{*}{HBB} & \multicolumn{2}{c|}{HOD-Net$^\dagger$ (Ours)} & 35.4 & \textbf{11.7} & \textbf{31.9} & 36.6 & 53.0 & \textbf{32.7} & 26.6 & 20.8 & \textbf{23.1} & \textbf{16.6} & 20.9 & \textbf{4.7} & \textbf{15.4} & 36.3 & ... & \textbf{53.2} \\ \hline
 & \multicolumn{2}{c|}{Oriented R-CNN$^\dagger$~\cite{xie2021oriented} (Resizing)} & 22.4 & 0.3 & 32.4 & 8.8 & 15.6 & 11.9 & 4.5 & 9.1 & 4.5 & 14.5 & 0.8 & 3.0 & 23.0 & 68.4 & ... & 30.9 \\
  & \multicolumn{2}{c|}{Oriented R-CNN$^\dagger$~\cite{xie2021oriented} (Cropping)} & 64.8 & 31.2 & \textbf{54.9} & \textbf{69.7} & 89.6 & 69.1 & 63.0 & 57.7 & 56.7 & 34.3 & 0.0 & 0.0 & 22.2 & 4.5 & ... & 41.9 \\
\multirow{-3}{*}{OBB} & \multicolumn{2}{c|}{HOD-Net$^\dagger$ (Ours)} & \textbf{65.2} & \textbf{32.6} & 53.4 & 66.4 & \textbf{89.7} & \textbf{69.3} & \textbf{63.1} & \textbf{59.3} & \textbf{57.6} & \textbf{38.7} & \textbf{35.7} & \textbf{46.9} & \textbf{29.4} & \textbf{78.6} & ... &\textbf{55.9} \\
\hline
	\end{tabular}}
	\label{table:object6}
 \vspace{-10pt}
\end{table*}

\begin{figure*}[!tb]
	\begin{center}
		\includegraphics[width=0.98\linewidth]{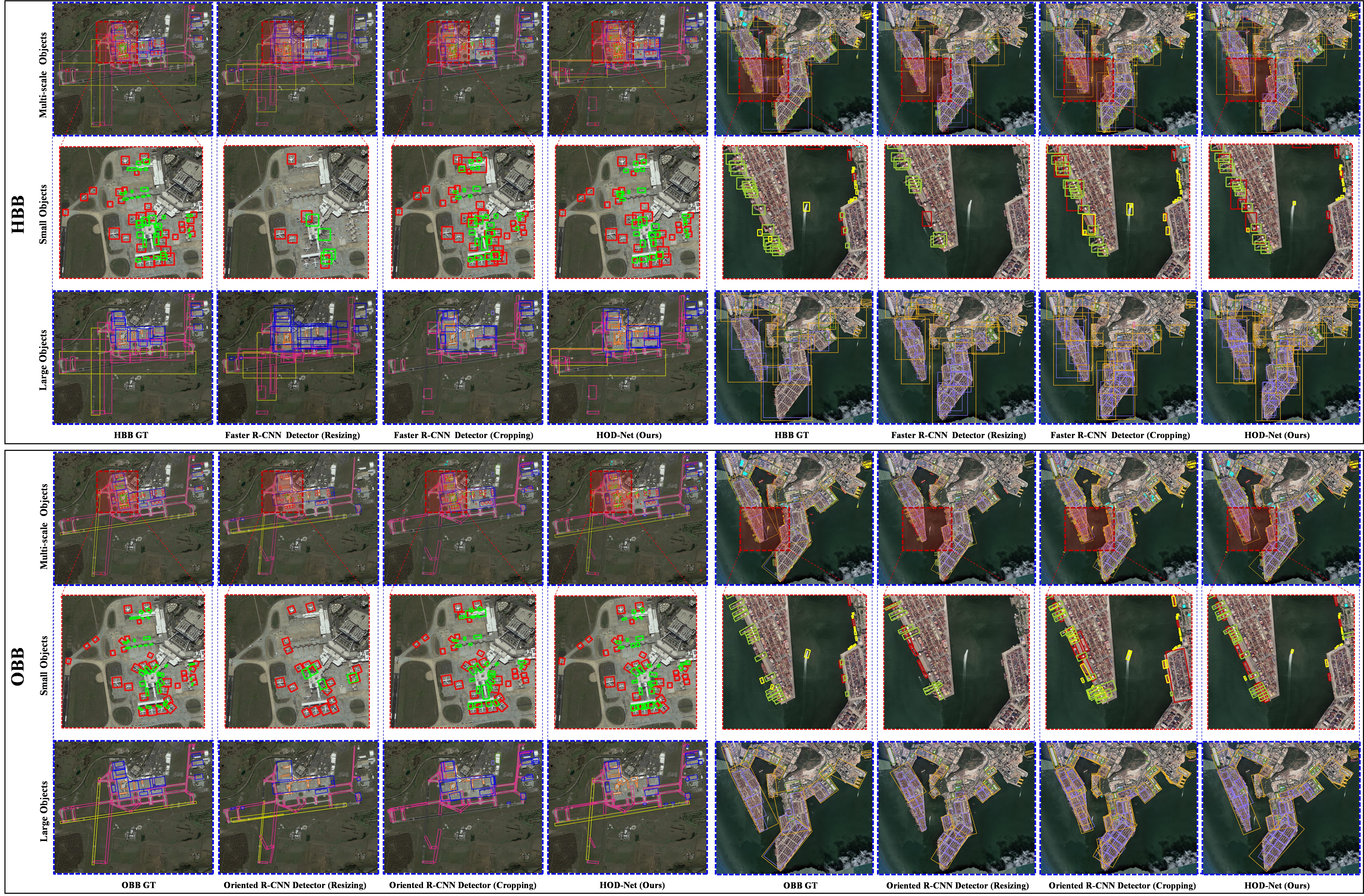}
	\end{center}
 \vspace{-15pt}
	\caption{The visualization results of OBB and HBB detection on the STAR dataset using the HOD-Net and comparison OBD methods.}
	\label{fig:object}
 \vspace{-5pt}
\end{figure*}

\section{Experiments and Analysis}	
\subsection{Benchmark}\label{subsec:Benchmark}

\textbf{Task Settings.} For the OBD task, we establish two types of OBD benchmarks: HBB-based and OBB-based detectors in large-size VHR SAI. For the SGG task, we evaluate SGG in large-size VHR SAI following three conventional sub-tasks of SGG in natural imagery: Predicate Classification (PredCls), Scene Graph Classification (SGCls) and Scene Graph Detection (SGDet). 




\textbf{Evaluation Metrics.} For the OBD task, mean average precision (mAP) is adopted as the evaluation metric. For the SGG task, different from the one-to-one relationship prediction in natural imagery, the relationship prediction for the SGG task in large-size VHR SAI is one-to-many. Based on the common evaluation metrics recall@K (R@K) and mean recall@K (mR@K) of the SGG task in natural imagery~\cite{tang2020unbiased}, we propose multi-label recall@K (MR@K) and mean multi-label recall@K (mMR@K) as the multi-label evaluation metrics for SGG in large-size VHR SAI.



To evaluate the performance of the SGG model more comprehensively, the harmonized mean (HMR@K) of both MR@K and mMR@K is taken as the comprehensive evaluation metric for the SGG task in large-size VHR SAI. For all tasks, the prediction can be considered as a true positive (TP)  result only when it matches the ground-truth triplet $<$subject, relationship, object$>$ and has at least 0.5 IoU between the bounding boxes of subject-object and the bounding boxes of ground-truth. As for the K, instead of the setting (K= 50/100), which is commonly used by the SGG task in natural imagery, we adopt the setting (K= 1,500/2,000) due to the average number of triplets per large-size VHR SAI far exceeds that in natural imagery.
\subsection{Implementation details}\label{subsec:details}
\textbf{Object Detector.} Taking the number of instances and the type of scenarios as the basis of division, we split the dataset into train set (60\%), val set (20\%), and test set (20\%). The detectors are trained on STAR train set using AdamW~\cite{loshchilov2018decoupled} as an default optimizer, and the performance is reported on STAR test set. All the listed models are trained on NVIDIA RTX3090 GPUs. We set the batch size to 2 and the initial learning rate to $1\times 10^{-4}$. The mean average precision (mAP) is calculated following PASCAL VOC 07~\cite{everingham2010pascal}. 


\textbf{Scene Graph Generation.} On top of the frozen detectors, we train all SGG models on STAR train set using SGD as an optimizer and evaluate them on STAR test set. Batch size and initial learning rate are 4 and $1\times 10^{-3}$ for the PredCls and SGCls tasks, and 2 and $1\times 10^{-3}$ for the SGDet task, respectively. For the SGDet task, object detection is performed using a Per-Class NMS~\cite{zellers2018neural,tang2020unbiased} with 0.5 IoU. Considering the computational resource consumption caused by numerous invalid object pairs in large-size VHR SAI, we use the proposed PPG network in the overall framework, which enables the SGG task in large-size VHR SAI to be realized in common computational conditions (\eg, one single GPU).

\begin{figure*}[!tb]
	\begin{center}
		\includegraphics[width=0.98\linewidth]{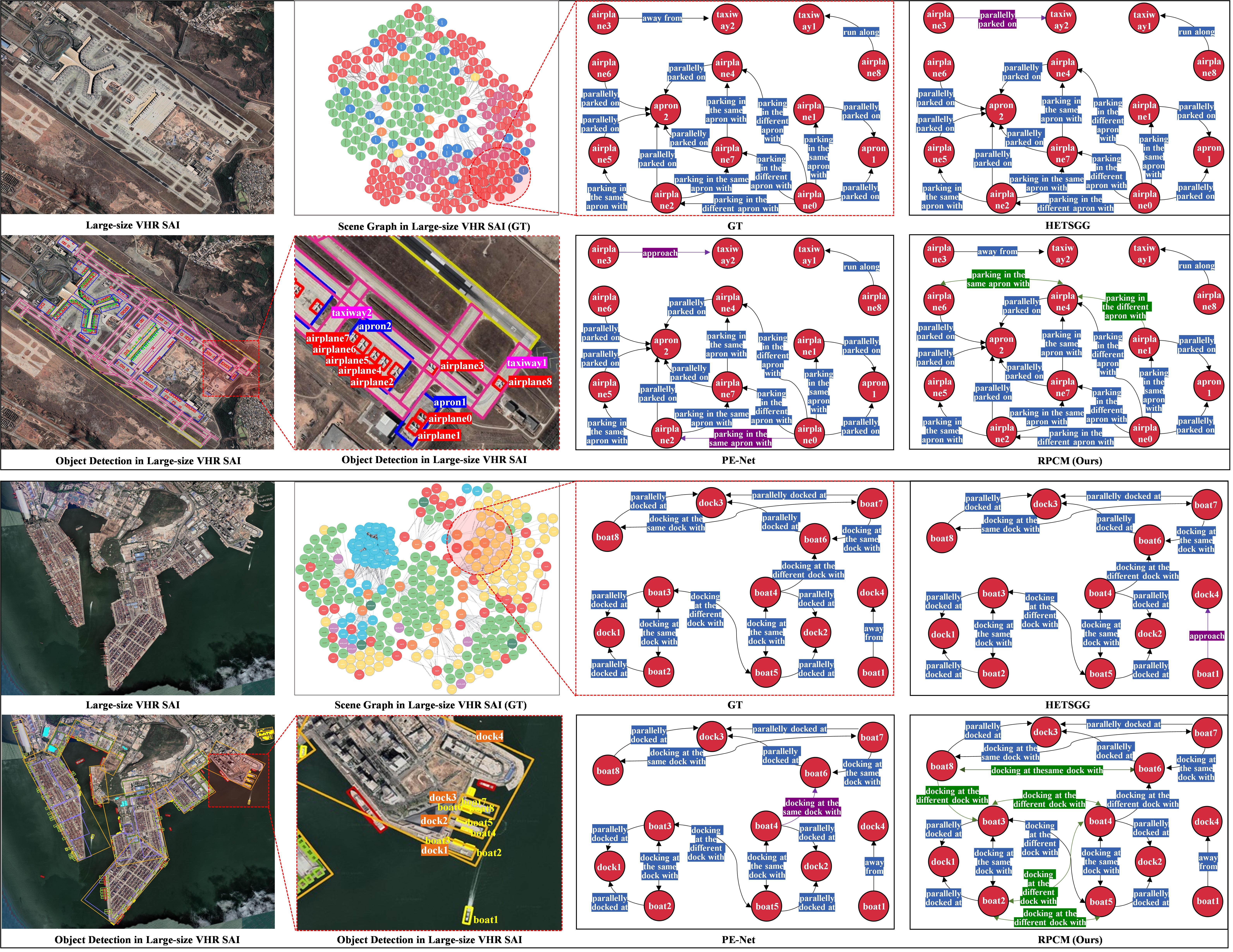}
	\end{center}
  \vspace{-15pt}
	\caption{The visualization results of the SGG task. Blue edges represent the correctly predicted relationships, purple edges indicate the ground-truth (GT) relationships that failed to be detected, and green edges are reasonable relationships predicted by the model but not annotated in the GT.}
	\label{fig:relationship}
\end{figure*}

\begin{table*}[tb!]
\renewcommand{\arraystretch}{2.0}
	\centering
	\caption{Baseline results (\%) of SGG on STAR test set for the PredCls, SGCls and SGDet tasks.}
 \vspace{-8pt}
	\resizebox{1.0\textwidth}{!}{
		\begin{tabular}{c|cc|ccc|ccc|ccc}
			\hline
 &  &  & \multicolumn{3}{c|}{\textbf{PredCls}} & \multicolumn{3}{c|}{\textbf{SGCls}} & \multicolumn{3}{c}{\textbf{SGDet}} \\ \cline{4-12} 
\multirow{-2}{*}{\begin{tabular}[c]{@{}c@{}}\textbf{OBD}\\ \textbf{Type}\end{tabular}} & \multirow{-2}{*}{\begin{tabular}[c]{@{}c@{}}\textbf{Relationship}\\ \textbf{Prediction}\end{tabular}} & \multirow{-2}{*}{\textbf{Venue}} & MR@1500/2000 & mMR@1500/2000 & HMR@1500/2000 & MR@1500/2000 & mMR@1500/2000 & HMR@1500/2000 & MR@1500/2000 & mMR@1500/2000 & HMR@1500/2000 \\ \hline
 & IMP~\cite{xu2017scene} & CVPR'17 & 49.90/51.62 & 18.51/19.40 & 27.00/28.20 & 42.48/43.63 & 16.46/17.02 & 23.73/24.49 & 23.01/23.72 & 8.25/8.60 & 12.15/12.62 \\
 & Motif~\cite{zellers2018neural} & CVPR'18 & 61.81/63.78 & 31.33/32.66 & 41.58/43.20 & 45.13/46.19 & 21.46/22.57 & 29.09/30.32 & 28.36/29.32 & 11.49/11.95 & 16.35/16.98 \\
 & GPS-Net~\cite{lin2020gps} & CVPR'20 & \textbf{65.68}/\textbf{67.10} & 34.92/36.11 & 45.60/46.95 & 31.99/33.16 & 14.61/15.40 & 20.06/21.03 & \textbf{30.83}/\textbf{31.58} & 12.71/13.10 & 18.00/18.52 \\
 & SHA~\cite{dong2022stacked} & CVPR'22 & 64.03/65.77 & 31.98/33.16 & 42.66/44.10 & \textbf{48.08}/\textbf{49.54} & 21.60/22.46 & 29.81/30.91 & 30.21/31.04 & 13.24/13.65 & 18.41/18.96 \\
 & HETSGG~\cite{yoon2023unbiased} & AAAI'23 & 63.83/65.26 & 28.70/29.72 & 39.60/40.84 & 30.03/31.08 & 10.50/11.05 & 15.56/16.30 & 25.19/25.82 & 8.17/8.50 & 12.34/12.79 \\
 & PE-Net~\cite{zheng2023prototype} & CVPR'23 & 63.21/65.11 & 35.94/37.10 & 45.82/47.27 & 38.84/40.05 & 19.46/20.20 & 25.93/26.86 & 28.11/29.22 & 12.80/13.40 & 17.59/18.37 \\
\multirow{-7}{*}{HBB} & \textbf{RPCM (Ours)} & - & 64.27/65.67 & \textbf{38.73}/\textbf{39.70} & \textbf{48.33}/\textbf{49.48} & 46.28/47.55 & \textbf{25.32}/\textbf{26.12} & \textbf{32.73}/\textbf{33.72} & 30.33/31.36 & \textbf{14.09}/\textbf{14.78} & \textbf{19.24}/\textbf{20.09} \\ \hline
 & IMP~\cite{xu2017scene} & CVPR'17 & 51.52/53.19 & 21.47/22.31 & 30.31/31.43 & 46.08/47.50 & 18.57/19.38 & 26.47/27.53 & 18.13/19.03 & 5.60/5.99 & 8.56/9.11 \\
 & Motif~\cite{zellers2018neural} & CVPR'18 & 62.02/64.00 & 29.40/30.64 & 39.89/41.44 & 50.44/52.29 & 24.21/25.36 & 32.72/34.16 & 20.63/21.60 & 7.78/8.22 & 11.30/11.91 \\
 & GPS-Net~\cite{lin2020gps} & CVPR'20 & \textbf{65.38}/66.87 & 32.52/34.08 & 44.80/46.22 & 29.17/30.36 & 14.28/14.90 & 19.17/19.99 & 24.18/25.04 & 9.54/10.09 & 13.68/14.38 \\
 & SHA~\cite{dong2022stacked} & CVPR'22 & 65.00/\textbf{67.04} & 33.26/34.78 & 44.00/45.80 & 51.01/52.63 & 23.30/24.15 & 31.99/33.11 & \textbf{28.01}/\textbf{29.06} & 10.18/10.79  & 14.93/15.74 \\
 & HETSGG~\cite{yoon2023unbiased} & AAAI'23 & 63.81/65.37 & 29.74/30.98 & 40.57/42.04 & 27.64/28.96 & 12.18/12.78 & 16.91/17.73 & 19.45/20.32 & 5.44/5.73 & 8.50/8.94 \\
 & PE-Net~\cite{zheng2023prototype} & CVPR'23 & 62.87/64.98 & 36.99/38.29 & 46.58/48.19 & 41.79/43.45 & 20.64/21.60 & 27.63/28.86 & 21.26/22.50 & 8.75/9.30 & 12.40/13.16 \\
\multirow{-7}{*}{OBB} & \textbf{RPCM (Ours)} & - & 64.23/65.86 & \textbf{41.24}/\textbf{42.30} & \textbf{50.23}/\textbf{51.51} & \textbf{51.29}/\textbf{52.72} & \textbf{30.04}/\textbf{30.85} & \textbf{37.89}/\textbf{38.92} & 27.23/28.50 & \textbf{11.53}/\textbf{12.07} & \textbf{16.20}/\textbf{16.96} \\
			\hline
	\end{tabular}}
	\label{table:relationship}
\end{table*}


\subsection{Results and Analysis for OBD}\label{subsec:RandAOBD}

\textbf{Comparison with Baselines.} To explore the properties of STAR and provide guidelines for future OBD task in large-size VHR SAI, we conduct a comprehensive evaluation of about 30 OBD methods and analyze the results. 
Table \ref{table:object30} provides a comprehensive benchmark on STAR, including single/two-stage, CNN/Transformer, anchor-based/anchor-free and supervised/weakly-supervised methods. We adopt MMDetection\footnote{\url{https://github.com/Zhuzi24/STAR-MMDetection}}~\cite{chen2019mmdetection} and MMRotate\footnote{\url{https://github.com/yangxue0827/STAR-MMRotate}}~\cite{zhou2022mmrotate} as the toolkits and all experiments are based on the standard `1x' (12 epochs) training schedule. It can be observed that our HOD-Net significantly outperforms other methods under the same backbone (ResNet50~\cite{he2016deep}) for both HBB-based and OBB-based detectors. To provide more accurate OBD results for the SGG task in large-size VHR SAI, we evaluate three OBD strategies on HBB-based and OBB-based detectors (see Table \ref{table:object6}): resizing, cropping, and our HOD-Net. It can be seen that our HOD-Net achieves 53.2\% and 55.9\% mAP on the HBB-based and OBB-based detectors, respectively (using a 0.5 IoU threshold). Specifically, under the same backbone (Swin-L~\cite{liu2021swin}), it achieves a 10.6\%/19.5\% improvement in mAP compared to the baseline Faster R-CNN detector (cropping/resizing) in the HBB detection task, and a 12.9\%/23.9\% improvement in mAP metric compared to the baseline Oriented R-CNN detector (cropping/resizing) in the OBB detection task. Notably, for objects with extreme aspect ratios (\eg, runways, ship\underline{ }locks), our HOD-Net outperforms other methods in OBB detection by a great margin of (34.9\%/31.2\%). Above experiments indicate that HOD-Net can provide more comprehensive and accurate object representation for SGG in large-size VHR SAI.


\textbf{Qualitative Results and Visualization.} In \figref{fig:object}, we further give the visualizations of different methods on OBB-based detectors and OBB-based detectors. It can be seen that many large objects (\eg, docks, aprons) cannot be detected holistically due to the integrity being destroyed in the detection based on cropping strategy, and a lot of small objects (\eg, airplanes, boats, cranes, boarding\underline{ }bridges) cannot be detected correctly due to the loss of information in the detection based on resizing strategy. The proposed HOD-Net shows superiority in the detection results of multi-scale objects and achieves almost the same results as the annotation for objects with extreme aspect ratios (\eg, runways). The above results show that HOD-Net can provide mandatory support for the SGG task in large-size VHR SAI.

\begin{figure*}[!tb]
	\begin{center}
		\includegraphics[width=0.98\linewidth]{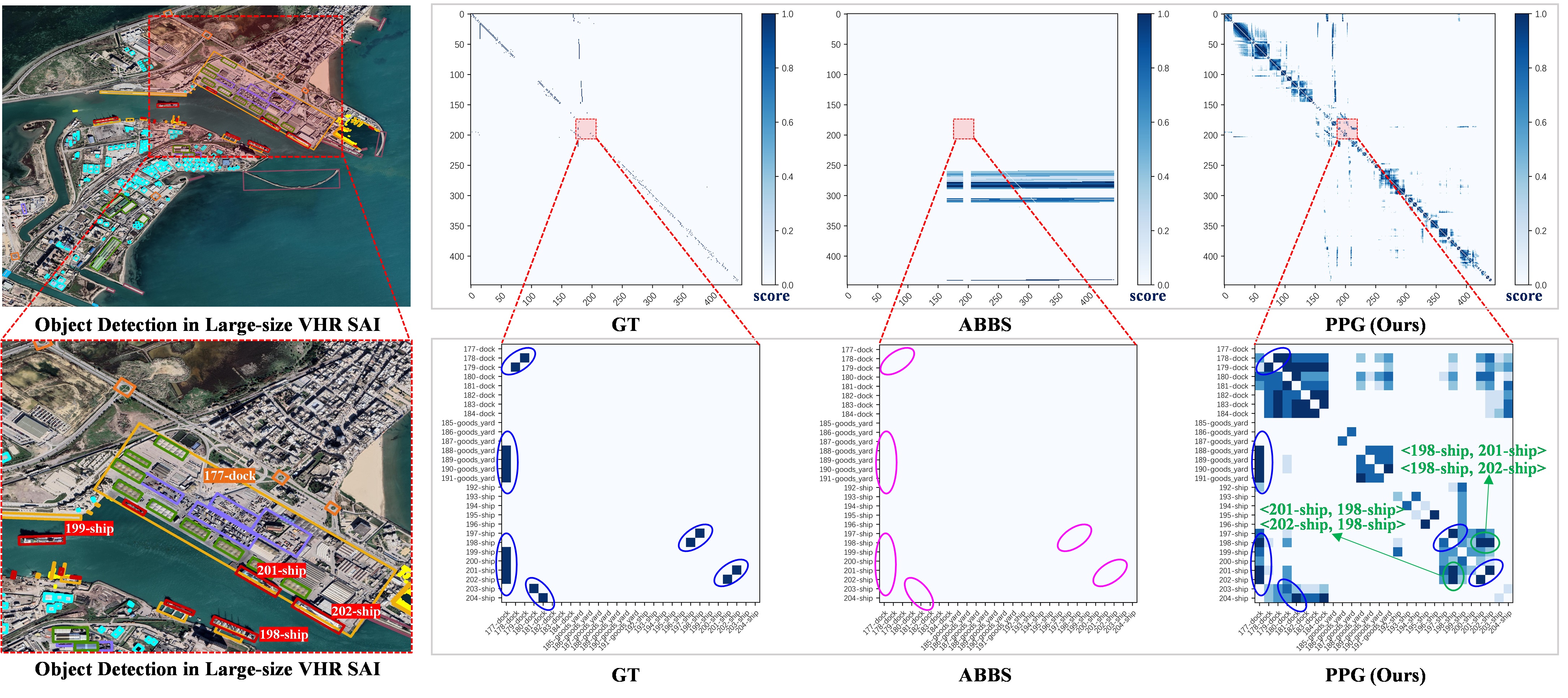}
	\end{center}
  \vspace{-15pt}
	\caption{The visualization results of pair score matrix. Blue edges represent the ground-truth (GT) pairs correctly predicted, purple edges indicate the GT pairs that failed to be detected, and green edges are reasonable pairs predicted by the model but not annotated in the GT.}
	\label{fig:pairs}
\end{figure*}

\begin{table*}[tb!]
\renewcommand{\arraystretch}{2.0}
	\centering
	\caption{Comparison of our PPG and other pair pruning methods on STAR test set for the PredCls, SGCls, and SGDet tasks.}
 \vspace{-8pt}
	\resizebox{1.0\textwidth}{!}{
		\begin{tabular}{c|ccc|ccc|ccc}
			\hline
 & \multicolumn{3}{c|}{\textbf{PredCls}} & \multicolumn{3}{c|}{\textbf{SGCls}} & \multicolumn{3}{c}{\textbf{SGDet}} \\ \cline{2-10} 
\multirow{-2}{*}{\textbf{Models}} & MR@1500/2000 & mMR@1500/2000 & HMR@1500/2000 & MR@1500/2000 & mMR@1500/2000 & HMR@1500/2000 & MR@1500/2000 & mMR@1500/2000 & HMR@1500/2000 \\ \hline
Randomization+RPCM & 51.49/52.62 & 31.96/32.46 & 39.44/40.15 & 43.12/43.54 & 26.02/26.28 & 32.46/32.78 & 16.31/16.57 & 6.66/6.76 & 9.46/9.60 \\
ABBS~\cite{li2024aug}+RPCM & 53.32/53.92 & 33.45/33.75 & 41.11/41.51 & 44.04/44.58 & 26.16/26.41 & 32.82/33.17 & 17.02/17.49 & 7.02/7.68 & 9.94/10.67 \\
\textbf{PPG+RPCM} (Ours) & \multicolumn{1}{c}{\textbf{64.23}/\textbf{65.86}} & \multicolumn{1}{c}{\textbf{41.24}/\textbf{42.30}} & \multicolumn{1}{c|}{\textbf{50.23}/\textbf{51.51}} & \textbf{51.29}/\textbf{52.72} &  \textbf{30.04}/\textbf{30.85} & \textbf{37.89}/\textbf{38.92} & \multicolumn{1}{c}{\textbf{27.23}/\textbf{28.50}} & \multicolumn{1}{c}{\textbf{11.53}/\textbf{12.07}} & \multicolumn{1}{c}{\textbf{16.20}/\textbf{16.96}} \\
			\hline
	\end{tabular}}
	\label{table:pairs}
\end{table*}

\subsection{Results and Analysis for SGG}\label{subsec:RandA}

\textbf{Comparison with the State-of-the-art Methods.} As shown in section 5.3, HOD-Net exhibits superior performance for the OBD task, which is a prerequisite for achieving accurate relationship prediction in the SGG task. For a fair comparison, all experiments utilized the same toolkit\footnote{\url{https://github.com/Zhuzi24/SGG-ToolKit}} and employed HOD-Net as the detector for the SGG task. We use MR@1500/2000, mMR@1500/2000, and HMR@1500/2000 as evaluation metrics of the SGG task. In Table \ref{table:relationship}, we conduct a comprehensive evaluation of 14 experiments and validated the effectiveness of RPCM by comparing it with six state-of-the-art SGG methods on both HBB-based and OBB-based detectors. As shown in Table \ref{table:relationship}, for the three sub-tasks of SGG, our RPCM shows better results overall, especially for the more challenging SGCls and SGDet tasks. For the SGCls task based on the OBB-based detector, our RPCM achieves 37.89\%/38.92\% on HMR@1500/2000, outperforming the SOTA method by a large margin of 5.17\%/4.76\%. For the SGDet task, our RPCM also outperforms the previous method with a 1.27\%/1.82\% improvement on HMR@1500/2000.


\textbf{Qualitative Results and Visualization.} The SGG visualization results of different models on the STAR dataset are shown in \figref{fig:relationship}. HETSGG has a limited ability to learn discriminative details(\eg, HETSGG fails to discriminate specific relationships $<$ship, away from/approach, dock$>$ by the wake trailing of the ship). PE-Net does not have the ability to constrain relationship prediction by relying on contextual reasoning (\eg, inferring the unreasonable triplet $<$airplane0, parking in the same apron with, airplane2$>$ via triplet $<$airplane2, parallelly parked on, apron2$>$ and triplet $<$airplane0, parallelly parked on, apron1$>$). In summary, our RPCM network can effectively guide the model in learning the discriminative details of different relationships.
\begin{table}[tb!]
\renewcommand{\arraystretch}{1.3}
	\centering
	\caption{Comparison with different PBA iters L.}
 \vspace{-8pt}
	\resizebox{0.45\textwidth}{!}{
		\begin{tabular}{cllccc}
			\hline
\multicolumn{3}{c}{} & \multicolumn{3}{c}{\textbf{PredCls}} \\ \cline{4-6} 
\multicolumn{3}{c}{\multirow{-2}{*}{\textbf{Iters L}}} & MR@1500/2000 & mMR@1500/2000 & HMR@1500/2000 \\ \hline
\multicolumn{3}{c}{L=1} & 63.56/65.44 & 39.12/40.54 & 48.43/50.06 \\
\multicolumn{3}{c}{L=2} & 64.14/65.79 & 40.10/41.18 & 49.35/50.65 \\
\multicolumn{3}{c}{L=3} & 63.90/65.54 & 40.79/41.88 & 49.79/51.10 \\
\multicolumn{3}{c}{\textbf{L=4}} & \textbf{64.23}/\textbf{65.86} & \textbf{41.24}/\textbf{42.30} & \textbf{50.23}/\textbf{51.51} \\
\multicolumn{3}{c}{L=5} & 62.14/64.09 & 39.92/41.09 & 48.61/50.08 \\ \hline
	  \end{tabular}}
	\label{table:iters}
\end{table}

\subsection{Ablation Study on SGG}\label{subsec:Ablation}

\subsubsection{Effect of PPG Network}\label{subsec:PPGNet}
Table \ref{table:pairs} gives the quantitative comparison of the proposed PPG network with other pair pruning methods on the STAR dataset when the sorted top 10,000 pairs are selected. It can be seen that the ABBS method based on statistics has only a weak improvement over the randomized method in all metrics, which shows that the semantic and spatial combination types of the object pairs in large-size VHR SAI are so complex that it is difficult to obtain combination patterns of objects by statistics. The score matrix, as depicted in the first row of \figref{fig:pairs}, becomes sparse after the PPG network learning, effectively retaining the annotated pairs. This highlights the robust learning capacity of the proposed PPG network when dealing with pairs containing rich knowledge. The local visualization shows that the PPG network can retain the annotated pairs with high scores, as well as some pairs that are non-annotated but may contain high-value relationships, this is because the annotation of triplets in the real annotation is hardly exhaustive, so achieving the same sparsity as ground-truth is not necessary.
\begin{table}[tb!]
\renewcommand{\arraystretch}{1.5}
	\centering
	\caption{Ablation study on each component of the RPCM. OCA and RCA denote object context augmentation and relationship context augmentation, respectively. Classifier indicates the method (prototype matching (PM)/ linear classification) for relationship prediction.}
 \vspace{-8pt}
	\resizebox{0.48\textwidth}{!}{
		\begin{tabular}{ccc|ccc}
			\hline
			\multicolumn{3}{c|}{\textbf{Component}} & \multicolumn{3}{c}{\textbf{PredCls}} \\ \hline
OCA & RCA & Classifier & MR@1500/2000 & mMR@1500/2000 & HMR@1500/2000 \\ \hline
\textbf{×} & \textbf{×} & PM & 63.10/65.28 & 36.53/38.05 & 46.27/48.08 \\
\textbf{\checkmark} & \textbf{×} & PM & 63.11/65.09 & 37.86/39.07 & \cellcolor[HTML]{FFFFFF}47.87/48.83 \\
\textbf{\checkmark} & \textbf{\checkmark} & \cellcolor[HTML]{FFFFFF}Linear & \textbf{66.55}/\textbf{68.18} & 34.58/35.80 & 45.51/46.95 \\
\textbf{\checkmark} & \textbf{\checkmark} & PM & 64.23/65.86 & \textbf{41.24}/\textbf{42.30} & \textbf{50.23}/\textbf{51.51} \\ \hline
	  \end{tabular}}
	\label{table:Ablation}
\end{table}
\begin{figure*}[!tb]
	\begin{center}
		\includegraphics[width=0.98\linewidth]{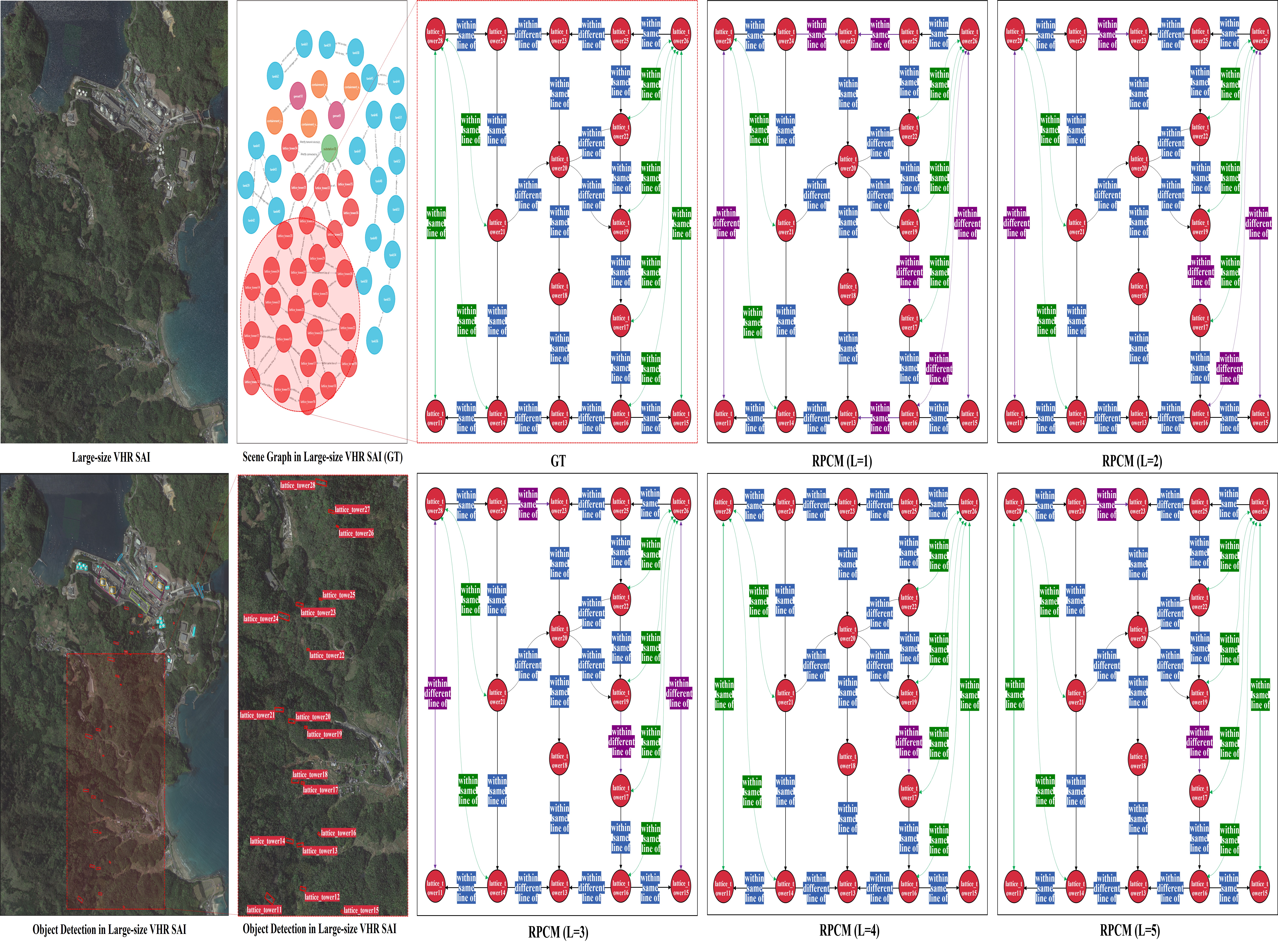}
	\end{center}
  \vspace{-15pt}
	\caption{Visualization of different PBA iters L. Blue edges represent the correctly predicted relationships, purple edges indicate incorrectly predicted relationships, and green edges are reasonable relationships predicted by the model but not annotated in the ground-truth (GT).}
	\label{fig:iters}
\end{figure*}
\subsubsection{Effect of PBA Iterations}\label{subsec:PBAIters}
In Table \ref{table:iters}, we compare the results on PredCls for different numbers of PBA iters L (L is set to 1, 2, 3, 4, 5). We find that the number of iters has little effect on our model. Notably, the HMR@K keeps improving with the increase in iters when iters are less than 4. This suggests that proper contextual information can effectively improve the model performance, but excessive introduction of contextual information may cause disturbance. In this paper, the number of iters for PBA is set to 4.

To explore the intrinsic impact of context introduction on the model, \figref{fig:iters} gives examples of visualizations obtained from different PBA iters. It can be seen that lattice\underline{ }towers that are far apart (\eg, "lattice\underline{ }tower11 and lattice\underline{ }tower28", "lattice\underline{ }tower15 and lattice\underline{ }tower26", and "lattice\underline{ }tower16 and lattice\underline{ }tower26") have incorrect relationship predictions due to insufficient contextual information when L is set to 1. As L increases, the introductions of more contextual information cause the wrong relationships to be gradually corrected. When L is set to 4, the relationships between lattice\underline{ }towers at different distances are predicted with better results. It is worth noting that the previous incorrect relationship prediction recurred when L is 5, which may be caused by excessive contextual interference. Overall, the introductions of proper contexts help relationship inference and excessive context brings disruption, consistent with the quantitative metrics in Table \ref{table:iters}.

\subsubsection{Effect of Different Components}\label{subsec:Comp}
We perform component ablation by gradually removing them from the RPCM on the PredCls task, as shown in Table \ref{table:Ablation}. Not surprisingly, the experimental results show that all three components, object context augmentation, relationship context augmentation, and prototype matching, are important for the performance of SGG. Specifically, prototype matching enables it to capture the discriminative features of different relationship categories in the semantic space, which in turn effectively distinguishes different categories with strong similarities. The object augmentation and relationship augmentation modules assist the model in learning contextual reasoning constraints for relationship prediction, as shown by the visualization in \figref{fig:relationship}.

\section{Conclusion}\label{sec:conclusion}
We have proposed a first-ever dataset named STAR for SGG in large-size VHR SAI. It covers 1,273 complex scenarios worldwide, with image sizes ranging from 512 × 768 to 27,860 × 31,096 pixels, including more than 210,000 objects (with both OBB and HBB annotations) and more than 400,000 triplet annotations. The large image size, the extensive sample volume, and the diversity of object scale and relationship semantics make STAR a valuable dataset, providing indispensable data support for advancing a new challenging and meaningful task: SGG in large-size VHR SAI. Furthermore, we propose the CAC framework, for the SGG task in large-size VHR SAI, to achieve a cascading understanding of SAI from three levels: object detection, pair pruning and relationship prediction. To facilitate the development of SGG in large VHR SAI, this paper releases an open-source toolkit containing various SGG methods and develops a large-scale benchmark in which the effectiveness of the proposed CAC framework is empirically validated. 

\section*{Acknowledgment}
This work was partly supported by the National Natural Science Foundation of China (42371321 and 42030102) and the Shanghai Municipal Science and Technology Major Project (2021SHZDZX0102).
\ifCLASSOPTIONcaptionsoff
  \newpage
\fi

\bibliographystyle{IEEEtran}
\bibliography{IEEEegbib}

\begin{thebibliography}{10}
\providecommand{\url}[1]{#1}
\csname url@samestyle\endcsname
\providecommand{\newblock}{\relax}
\providecommand{\bibinfo}[2]{#2}
\providecommand{\BIBentrySTDinterwordspacing}{\spaceskip=0pt\relax}
\providecommand{\BIBentryALTinterwordstretchfactor}{4}
\providecommand{\BIBentryALTinterwordspacing}{\spaceskip=\fontdimen2\font plus
\BIBentryALTinterwordstretchfactor\fontdimen3\font minus \fontdimen4\font\relax}
\providecommand{\BIBforeignlanguage}[2]{{%
\expandafter\ifx\csname l@#1\endcsname\relax
\typeout{** WARNING: IEEEtran.bst: No hyphenation pattern has been}%
\typeout{** loaded for the language `#1'. Using the pattern for}%
\typeout{** the default language instead.}%
\else
\language=\csname l@#1\endcsname
\fi
#2}}
\providecommand{\BIBdecl}{\relax}
\BIBdecl

\bibitem{chang2021comprehensive}
X.~Chang, P.~Ren, P.~Xu, Z.~Li, X.~Chen, and A.~Hauptmann, ``A comprehensive survey of scene graphs: Generation and application,'' \emph{IEEE Transactions on Pattern Analysis and Machine Intelligence}, vol.~45, no.~1, pp. 1--26, 2021.

\bibitem{cong2023reltr}
Y.~Cong, M.~Y. Yang, and B.~Rosenhahn, ``Reltr: Relation transformer for scene graph generation,'' \emph{IEEE Transactions on Pattern Analysis and Machine Intelligence}, vol.~45, no.~9, pp. 11\,169--11\,183, 2023.

\bibitem{Sun2023Unbiased}
S.~Sun, S.~Zhi, Q.~Liao, J.~Heikkilä, and L.~Liu, ``Unbiased scene graph generation via two-stage causal modeling,'' \emph{IEEE Transactions on Pattern Analysis and Machine Intelligence}, vol.~45, no.~10, pp. 12\,562--12\,580, 2023.

\bibitem{xia2018dota}
G.-S. Xia, X.~Bai, J.~Ding, Z.~Zhu, S.~Belongie, J.~Luo, M.~Datcu, M.~Pelillo, and L.~Zhang, ``Dota: A large-scale dataset for object detection in aerial images,'' in \emph{Proceedings of the IEEE/CVF Conference on Computer Vision and Pattern Recognition}, 2018, pp. 3974--3983.

\bibitem{sun2022fair1m}
X.~Sun, P.~Wang, Z.~Yan, F.~Xu, R.~Wang, W.~Diao, J.~Chen, J.~Li, Y.~Feng, T.~Xu \emph{et~al.}, ``Fair1m: A benchmark dataset for fine-grained object recognition in high-resolution remote sensing imagery,'' \emph{ISPRS Journal of Photogrammetry and Remote Sensing}, vol. 184, pp. 116--130, 2022.

\bibitem{yang2022scrdet++}
X.~Yang, J.~Yan, W.~Liao, X.~Yang, J.~Tang, and T.~He, ``Scrdet++: Detecting small, cluttered and rotated objects via instance-level feature denoising and rotation loss smoothing,'' \emph{IEEE Transactions on Pattern Analysis and Machine Intelligence}, vol.~45, no.~2, pp. 2384--2399, 2022.

\bibitem{cheng2017remote}
G.~Cheng, J.~Han, and X.~Lu, ``Remote sensing image scene classification: Benchmark and state of the art,'' \emph{Proceedings of the IEEE}, vol. 105, no.~10, pp. 1865--1883, 2017.

\bibitem{nogueira2017towards}
K.~Nogueira, O.~A. Penatti, and J.~A. Dos~Santos, ``Towards better exploiting convolutional neural networks for remote sensing scene classification,'' \emph{Pattern Recognition}, vol.~61, pp. 539--556, 2017.

\bibitem{zhu2021spatial}
H.~Zhu, M.~Ma, W.~Ma, L.~Jiao, S.~Hong, J.~Shen, and B.~Hou, ``A spatial-channel progressive fusion resnet for remote sensing classification,'' \emph{Information Fusion}, vol.~70, pp. 72--87, 2021.

\bibitem{hossain2019segmentation}
M.~D. Hossain and D.~Chen, ``Segmentation for object-based image analysis (obia): A review of algorithms and challenges from remote sensing perspective,'' \emph{ISPRS Journal of Photogrammetry and Remote Sensing}, vol. 150, pp. 115--134, 2019.

\bibitem{wurm2019semantic}
M.~Wurm, T.~Stark, X.~X. Zhu, M.~Weigand, and H.~Taubenb{\"o}ck, ``Semantic segmentation of slums in satellite images using transfer learning on fully convolutional neural networks,'' \emph{ISPRS Journal of Photogrammetry and Remote Sensing}, vol. 150, pp. 59--69, 2019.

\bibitem{li2021learning}
Y.~Li, T.~Shi, Y.~Zhang, W.~Chen, Z.~Wang, and H.~Li, ``Learning deep semantic segmentation network under multiple weakly-supervised constraints for cross-domain remote sensing image semantic segmentation,'' \emph{ISPRS Journal of Photogrammetry and Remote Sensing}, vol. 175, pp. 20--33, 2021.

\bibitem{johnson2015image}
J.~Johnson, R.~Krishna, M.~Stark, L.-J. Li, D.~Shamma, M.~Bernstein, and L.~Fei-Fei, ``Image retrieval using scene graphs,'' in \emph{Proceedings of the IEEE/CVF Conference on Computer Vision and Pattern Recognition}, 2015, pp. 3668--3678.

\bibitem{yang2019auto}
X.~Yang, K.~Tang, H.~Zhang, and J.~Cai, ``Auto-encoding scene graphs for image captioning,'' in \emph{Proceedings of the IEEE/CVF Conference on Computer Vision and Pattern Recognition}, 2019, pp. 10\,685--10\,694.

\bibitem{Li_2019_ICCV}
L.~Li, Z.~Gan, Y.~Cheng, and J.~Liu, ``Relation-aware graph attention network for visual question answering,'' in \emph{Proceedings of the IEEE/CVF International Conference on Computer Vision}, October 2019.

\bibitem{reinartz2006traffic}
P.~Reinartz, M.~Lachaise, E.~Schmeer, T.~Krauss, and H.~Runge, ``Traffic monitoring with serial images from airborne cameras,'' \emph{ISPRS Journal of Photogrammetry and Remote Sensing}, vol.~61, no. 3-4, pp. 149--158, 2006.

\bibitem{smigaj2023monitoring}
M.~Smigaj, C.~R. Hackney, P.~K. Diem, N.~T. Ngoc, D.~Du~Bui, S.~E. Darby, J.~Leyland \emph{et~al.}, ``Monitoring riverine traffic from space: The untapped potential of remote sensing for measuring human footprint on inland waterways,'' \emph{Science of the Total Environment}, vol. 860, p. 160363, 2023.

\bibitem{matikainen2016remote}
L.~Matikainen, M.~Lehtom{\"a}ki, E.~Ahokas, J.~Hyypp{\"a}, M.~Karjalainen, A.~Jaakkola, A.~Kukko, and T.~Heinonen, ``Remote sensing methods for power line corridor surveys,'' \emph{ISPRS Journal of Photogrammetry and Remote sensing}, vol. 119, pp. 10--31, 2016.

\bibitem{chen2021message}
J.~Chen, X.~Zhou, Y.~Zhang, G.~Sun, M.~Deng, and H.~Li, ``Message-passing-driven triplet representation for geo-object relational inference in hrsi,'' \emph{IEEE Geoscience and Remote Sensing Letters}, vol.~19, pp. 1--5, 2021.

\bibitem{li2021semantic}
P.~Li, D.~Zhang, A.~Wulamu, X.~Liu, and P.~Chen, ``Semantic relation model and dataset for remote sensing scene understanding,'' \emph{ISPRS International Journal of Geo-Information}, vol.~10, no.~7, p. 488, 2021.

\bibitem{li2020object}
K.~Li, G.~Wan, G.~Cheng, L.~Meng, and J.~Han, ``Object detection in optical remote sensing images: A survey and a new benchmark,'' \emph{ISPRS Journal of Photogrammetry and Remote Sensing}, vol. 159, pp. 296--307, 2020.

\bibitem{ding2021object}
J.~Ding, N.~Xue, G.-S. Xia, X.~Bai, W.~Yang, M.~Y. Yang, S.~Belongie, J.~Luo, M.~Datcu, M.~Pelillo \emph{et~al.}, ``Object detection in aerial images: A large-scale benchmark and challenges,'' \emph{IEEE Transactions on Pattern Analysis and Machine Intelligence}, 2021.

\bibitem{cheng2016learning}
G.~Cheng, P.~Zhou, and J.~Han, ``Learning rotation-invariant convolutional neural networks for object detection in vhr optical remote sensing images,'' \emph{IEEE Transactions on Geoscience and Remote Sensing}, vol.~54, no.~12, pp. 7405--7415, 2016.

\bibitem{zhang2019hierarchical}
Y.~Zhang, Y.~Yuan, Y.~Feng, and X.~Lu, ``Hierarchical and robust convolutional neural network for very high-resolution remote sensing object detection,'' \emph{IEEE Transactions on Geoscience and Remote Sensing}, vol.~57, no.~8, pp. 5535--5548, 2019.

\bibitem{cheng2022anchor}
G.~Cheng, J.~Wang, K.~Li, X.~Xie, C.~Lang, Y.~Yao, and J.~Han, ``Anchor-free oriented proposal generator for object detection,'' \emph{IEEE Transactions on Geoscience and Remote Sensing}, 2022.

\bibitem{li2024learning}
Y.~Li, J.~Luo, Y.~Zhang, Y.~Tan, J.-G. Yu, and S.~Bai, ``Learning to holistically detect bridges from large-size vhr remote sensing imagery,'' \emph{IEEE Transactions on Pattern Analysis and Machine Intelligence}, vol.~44, no.~11, pp. 7778--7796, 2024.

\bibitem{schuster2015generating}
S.~Schuster, R.~Krishna, A.~Chang, L.~Fei-Fei, and C.~D. Manning, ``Generating semantically precise scene graphs from textual descriptions for improved image retrieval,'' in \emph{Proceedings of the Fourth Workshop on Vision and Language}, 2015, pp. 70--80.

\bibitem{Kim_2019_CVPR}
D.-J. Kim, J.~Choi, T.-H. Oh, and I.~S. Kweon, ``Dense relational captioning: Triple-stream networks for relationship-based captioning,'' in \emph{Proceedings of the IEEE/CVF Conference on Computer Vision and Pattern Recognition}, 2019.

\bibitem{sadeghi2011recognition}
M.~A. Sadeghi and A.~Farhadi, ``Recognition using visual phrases,'' in \emph{Proceedings of the IEEE/CVF Conference on Computer Vision and Pattern Recognition}, 2011, pp. 1745--1752.

\bibitem{thomee2015new}
B.~Thomee, D.~A. Shamma, G.~Friedland, B.~Elizalde, K.~Ni, D.~Poland, D.~Borth, and L.-J. Li, ``The new data and new challenges in multimedia research,'' \emph{arXiv preprint arXiv:1503.01817}, vol.~1, no.~8, 2015.

\bibitem{lin2014microsoft}
T.-Y. Lin, M.~Maire, S.~Belongie, J.~Hays, P.~Perona, D.~Ramanan, P.~Doll{\'a}r, and C.~L. Zitnick, ``Microsoft coco: Common objects in context,'' in \emph{Proceedings of the European Conference on Computer Vision}.\hskip 1em plus 0.5em minus 0.4em\relax Springer, 2014, pp. 740--755.

\bibitem{lu2016visual}
C.~Lu, R.~Krishna, M.~Bernstein, and L.~Fei-Fei, ``Visual relationship detection with language priors,'' in \emph{Proceedings of the European Conference on Computer Vision}, 2016, pp. 852--869.

\bibitem{krishna2017visual}
R.~Krishna, Y.~Zhu, O.~Groth, J.~Johnson, K.~Hata, J.~Kravitz, S.~Chen, Y.~Kalantidis, L.-J. Li, D.~A. Shamma \emph{et~al.}, ``Visual genome: Connecting language and vision using crowdsourced dense image annotations,'' \emph{International Journal of Computer Vision}, vol. 123, pp. 32--73, 2017.

\bibitem{xu2017scene}
D.~Xu, Y.~Zhu, C.~B. Choy, and L.~Fei-Fei, ``Scene graph generation by iterative message passing,'' in \emph{Proceedings of the IEEE/CVF Conference on Computer Vision and Pattern Recognition}, 2017, pp. 5410--5419.

\bibitem{liang2019vrr}
Y.~Liang, Y.~Bai, W.~Zhang, X.~Qian, L.~Zhu, and T.~Mei, ``Vrr-vg: Refocusing visually-relevant relationships,'' in \emph{Proceedings of the IEEE/CVF International Conference on Computer Vision}, 2019, pp. 10\,403--10\,412.

\bibitem{hudson2019gqa}
D.~A. Hudson and C.~D. Manning, ``Gqa: A new dataset for real-world visual reasoning and compositional question answering,'' in \emph{Proceedings of the IEEE/CVF Conference on Computer Vision and Pattern Recognition}, 2019, pp. 6700--6709.

\bibitem{kuznetsova2020open}
A.~Kuznetsova, H.~Rom, N.~Alldrin, J.~Uijlings, I.~Krasin, J.~Pont-Tuset, S.~Kamali, S.~Popov, M.~Malloci, A.~Kolesnikov \emph{et~al.}, ``The open images dataset v4: Unified image classification, object detection, and visual relationship detection at scale,'' \emph{International Journal of Computer Vision}, vol. 128, no.~7, pp. 1956--1981, 2020.

\bibitem{lin2022srsg}
Z.~Lin, F.~Zhu, Y.~Kong, Q.~Wang, and J.~Wang, ``Srsg and s2sg: a model and a dataset for scene graph generation of remote sensing images from segmentation results,'' \emph{IEEE Transactions on Geoscience and Remote Sensing}, vol.~60, pp. 1--11, 2022.

\bibitem{li2024aug}
Y.~Li, K.~Li, Y.~Zhang, L.~Wang, and D.~Zhang, ``Aug: A new dataset and an efficient model for aerial image urban scene graph generation,'' \emph{arXiv preprint arXiv:2404.07788}, 2024.

\bibitem{lu2017exploring}
X.~Lu, B.~Wang, X.~Zheng, and X.~Li, ``Exploring models and data for remote sensing image caption generation,'' \emph{IEEE Transactions on Geoscience and Remote Sensing}, vol.~56, no.~4, pp. 2183--2195, 2017.

\bibitem{tang2020unbiased}
K.~Tang, Y.~Niu, J.~Huang, J.~Shi, and H.~Zhang, ``Unbiased scene graph generation from biased training,'' in \emph{Proceedings of the IEEE/CVF Conference on Computer Vision and Pattern Recognition}, 2020, pp. 3716--3725.

\bibitem{dong2022stacked}
X.~Dong, T.~Gan, X.~Song, J.~Wu, Y.~Cheng, and L.~Nie, ``Stacked hybrid-attention and group collaborative learning for unbiased scene graph generation,'' in \emph{Proceedings of the IEEE/CVF Conference on Computer Vision and Pattern Recognition}, 2022, pp. 19\,427--19\,436.

\bibitem{zellers2018neural}
R.~Zellers, M.~Yatskar, S.~Thomson, and Y.~Choi, ``Neural motifs: Scene graph parsing with global context,'' in \emph{Proceedings of the IEEE Conference on Computer Vision and Pattern Recognition}, 2018, pp. 5831--5840.

\bibitem{tang2019learning}
K.~Tang, H.~Zhang, B.~Wu, W.~Luo, and W.~Liu, ``Learning to compose dynamic tree structures for visual contexts,'' in \emph{Proceedings of the IEEE/CVF Conference on Computer Vision and Pattern Recognition}, 2019, pp. 6619--6628.

\bibitem{zheng2023prototype}
C.~Zheng, X.~Lyu, L.~Gao, B.~Dai, and J.~Song, ``Prototype-based embedding network for scene graph generation,'' in \emph{Proceedings of the IEEE/CVF Conference on Computer Vision and Pattern Recognition}, 2023, pp. 22\,783--22\,792.

\bibitem{yoon2023unbiased}
K.~Yoon, K.~Kim, J.~Moon, and C.~Park, ``Unbiased heterogeneous scene graph generation with relation-aware message passing neural network,'' in \emph{Proceedings of the AAAI Conference on Artificial Intelligence}, vol.~37, no.~3, 2023, pp. 3285--3294.

\bibitem{chen2022resistance}
C.~Chen, Y.~Zhan, B.~Yu, L.~Liu, Y.~Luo, and B.~Du, ``Resistance training using prior bias: toward unbiased scene graph generation,'' in \emph{Proceedings of the AAAI Conference on Artificial Intelligence}, vol.~36, no.~1, 2022, pp. 212--220.

\bibitem{deng2022hierarchical}
Y.~Deng, Y.~Li, Y.~Zhang, X.~Xiang, J.~Wang, J.~Chen, and J.~Ma, ``Hierarchical memory learning for fine-grained scene graph generation,'' in \emph{Proceedings of the European Conference on Computer Vision}, 2022, pp. 266--283.

\bibitem{yang2018graph}
J.~Yang, J.~Lu, S.~Lee, D.~Batra, and D.~Parikh, ``Graph r-cnn for scene graph generation,'' in \emph{Proceedings of the European Conference on Computer Vision}, 2018, pp. 670--685.

\bibitem{lin2020gps}
X.~Lin, C.~Ding, J.~Zeng, and D.~Tao, ``Gps-net: Graph property sensing network for scene graph generation,'' in \emph{Proceedings of the IEEE/CVF Conference on Computer Vision and Pattern Recognition}, 2020, pp. 3746--3753.

\bibitem{li2021bipartite}
R.~Li, S.~Zhang, B.~Wan, and X.~He, ``Bipartite graph network with adaptive message passing for unbiased scene graph generation,'' in \emph{Proceedings of the IEEE/CVF Conference on Computer Vision and Pattern Recognition}, 2021, pp. 11\,109--11\,119.

\bibitem{shi2017can}
Z.~Shi and Z.~Zou, ``Can a machine generate humanlike language descriptions for a remote sensing image?'' \emph{IEEE Transactions on Geoscience and Remote Sensing}, vol.~55, no.~6, pp. 3623--3634, 2017.

\bibitem{yang2018automatic}
X.~Yang, H.~Sun, K.~Fu, J.~Yang, X.~Sun, M.~Yan, and Z.~Guo, ``Automatic ship detection in remote sensing images from google earth of complex scenes based on multiscale rotation dense feature pyramid networks,'' \emph{Remote sensing}, vol.~10, no.~1, p. 132, 2018.

\bibitem{ding2019learning}
J.~Ding, N.~Xue, Y.~Long, G.-S. Xia, and Q.~Lu, ``Learning roi transformer for oriented object detection in aerial images,'' in \emph{Proceedings of the IEEE/CVF conference on computer vision and pattern recognition}, 2019, pp. 2849--2858.

\bibitem{yang2019scrdet}
X.~Yang, J.~Yang, J.~Yan, Y.~Zhang, T.~Zhang, Z.~Guo, X.~Sun, and K.~Fu, ``Scrdet: Towards more robust detection for small, cluttered and rotated objects,'' in \emph{Proceedings of the IEEE/CVF international conference on computer vision}, 2019, pp. 8232--8241.

\bibitem{hou2023g}
L.~Hou, K.~Lu, X.~Yang, Y.~Li, and J.~Xue, ``G-rep: Gaussian representation for arbitrary-oriented object detection,'' \emph{Remote Sensing}, vol.~15, no.~3, p. 757, 2023.

\bibitem{yang2023h2rbox}
X.~Yang, G.~Zhang, W.~Li, X.~Wang, Y.~Zhou, and J.~Yan, ``H2rbox: Horizontal box annotation is all you need for oriented object detection,'' \emph{International Conference on Learning Representations}, 2023.

\bibitem{yu2023h2rboxv2}
Y.~Yu, X.~Yang, Q.~Li, Y.~Zhou, G.~Zhang, F.~Da, and J.~Yan, ``H2rbox-v2: Incorporating symmetry for boosting horizontal box supervised oriented object detection,'' in \emph{Advances in Neural Information Processing Systems}, 2023.

\bibitem{luo2024pointobb}
J.~Luo, X.~Yang, Y.~Yu, Q.~Li, J.~Yan, and Y.~Li, ``Pointobb: Learning oriented object detection via single point supervision,'' in \emph{IEEE/CVF Conference on Computer Vision and Pattern Recognition}, 2024.

\bibitem{yu2024point2rbox}
Y.~Yu, X.~Yang, Q.~Li, F.~Da, J.~Dai, Y.~Qiao, and J.~Yan, ``Point2rbox: Combine knowledge from synthetic visual patterns for end-to-end oriented object detection with single point supervision,'' in \emph{IEEE/CVF Conference on Computer Vision and Pattern Recognition}, 2024.

\bibitem{akyon2022slicing}
F.~C. Akyon, S.~O. Altinuc, and A.~Temizel, ``Slicing aided hyper inference and fine-tuning for small object detection,'' in \emph{Proceedings of the IEEE International Conference on Image Processing}, 2022, pp. 966--970.

\bibitem{chen2023coupled}
X.~Chen, C.~Wang, Z.~Li, M.~Liu, Q.~Li, H.~Qi, D.~Ma, Z.~Li, and Y.~Wang, ``Coupled global--local object detection for large vhr aerial images,'' \emph{Knowledge-Based Systems}, vol. 260, p. 110097, 2023.

\bibitem{girshick2015fast}
R.~Girshick, ``Fast r-cnn,'' in \emph{Proceedings of the IEEE/CVF International Conference on Computer Vision}, 2015, pp. 1440--1448.

\bibitem{goodfellow2014generative}
I.~Goodfellow, J.~Pouget-Abadie, M.~Mirza, B.~Xu, D.~Warde-Farley, S.~Ozair, A.~Courville, and Y.~Bengio, ``Generative adversarial nets,'' \emph{Advances in Neural Information Processing Systems}, vol.~27, 2014.

\bibitem{velickovic2018graph}
P.~Casanova, A.~R.~P. Lio, and Y.~Bengio, ``Graph attention networks,'' \emph{ICLR. Petar Velickovic Guillem Cucurull Arantxa Casanova Adriana Romero Pietro Li{\`o} and Yoshua Bengio}, 2018.

\bibitem{nair2010rectified}
V.~Nair and G.~E. Hinton, ``Rectified linear units improve restricted boltzmann machines,'' in \emph{Proceedings of the 27th International Conference on Machine Learning}, 2010, pp. 807--814.

\bibitem{liu2021swin}
Z.~Liu, Y.~Lin, Y.~Cao, H.~Hu, Y.~Wei, Z.~Zhang, S.~Lin, and B.~Guo, ``Swin transformer: Hierarchical vision transformer using shifted windows,'' in \emph{Proceedings of the IEEE/CVF International Conference on Computer Vision}, 2021, pp. 10\,012--10\,022.

\bibitem{he2016deep}
K.~He, X.~Zhang, S.~Ren, and J.~Sun, ``Deep residual learning for image recognition,'' in \emph{Proceedings of the IEEE Conference on Computer Vision and Pattern Recognition}, 2016, pp. 770--778.

\bibitem{ren2015faster}
S.~Ren, K.~He, R.~Girshick, and J.~Sun, ``Faster r-cnn: Towards real-time object detection with region proposal networks,'' \emph{Advances in Neural Information Processing Systems}, vol.~28, 2015.

\bibitem{lin2017focal}
T.-Y. Lin, P.~Goyal, R.~Girshick, K.~He, and P.~Doll{\'a}r, ``Focal loss for dense object detection,'' in \emph{Proceedings of the IEEE/CVF International Conference on Computer Vision}, 2017, pp. 2980--2988.

\bibitem{cai2019cascade}
Z.~Cai and N.~Vasconcelos, ``Cascade r-cnn: High quality object detection and instance segmentation,'' \emph{IEEE transactions on pattern analysis and machine intelligence}, vol.~43, no.~5, pp. 1483--1498, 2019.

\bibitem{tian2020fcos}
Z.~Tian, C.~Shen, H.~Chen, and T.~He, ``Fcos: A simple and strong anchor-free object detector,'' \emph{IEEE Transactions on Pattern Analysis and Machine Intelligence}, vol.~44, no.~4, pp. 1922--1933, 2020.

\bibitem{feng2021tood}
C.~Feng, Y.~Zhong, Y.~Gao, M.~R. Scott, and W.~Huang, ``Tood: Task-aligned one-stage object detection,'' in \emph{2021 IEEE/CVF International Conference on Computer Vision (ICCV)}.\hskip 1em plus 0.5em minus 0.4em\relax IEEE Computer Society, 2021, pp. 3490--3499.

\bibitem{zhu2020deformable}
X.~Zhu, W.~Su, L.~Lu, B.~Li, X.~Wang, and J.~Dai, ``Deformable detr: Deformable transformers for end-to-end object detection,'' in \emph{International Conference on Learning Representations}, 2020.

\bibitem{zeng2024ars}
Y.~Zeng, Y.~Chen, X.~Yang, Q.~Li, and J.~Yan, ``Ars-detr: Aspect ratio-sensitive detection transformer for aerial oriented object detection,'' \emph{IEEE Transactions on Geoscience and Remote Sensing}, vol.~62, pp. 1--15, 2024.

\bibitem{zhang2020bridging}
S.~Zhang, C.~Chi, Y.~Yao, Z.~Lei, and S.~Z. Li, ``Bridging the gap between anchor-based and anchor-free detection via adaptive training sample selection,'' in \emph{Proceedings of the IEEE/CVF conference on computer vision and pattern recognition}, 2020, pp. 9759--9768.

\bibitem{yang2021learning}
X.~Yang, X.~Yang, J.~Yang, Q.~Ming, W.~Wang, Q.~Tian, and J.~Yan, ``Learning high-precision bounding box for rotated object detection via kullback-leibler divergence,'' \emph{Advances in Neural Information Processing Systems}, vol.~34, pp. 18\,381--18\,394, 2021.

\bibitem{yang2021rethinking}
X.~Yang, J.~Yan, Q.~Ming, W.~Wang, X.~Zhang, and Q.~Tian, ``Rethinking rotated object detection with gaussian wasserstein distance loss,'' in \emph{International Conference on Machine Learning}.\hskip 1em plus 0.5em minus 0.4em\relax PMLR, 2021, pp. 11\,830--11\,841.

\bibitem{yang2023kfiou}
X.~Yang, Y.~Zhou, G.~Zhang, J.~Yang, W.~Wang, J.~Yan, X.~Zhang, and Q.~Tian, ``The kfiou loss for rotated object detection,'' in \emph{International Conference on Learning Representations}, 2023.

\bibitem{xu2023dynamic}
C.~Xu, J.~Ding, J.~Wang, W.~Yang, H.~Yu, L.~Yu, and G.-S. Xia, ``Dynamic coarse-to-fine learning for oriented tiny object detection,'' in \emph{Proceedings of the IEEE/CVF Conference on Computer Vision and Pattern Recognition}, 2023, pp. 7318--7328.

\bibitem{yang2021r3det}
X.~Yang, J.~Yan, Z.~Feng, and T.~He, ``R3det: Refined single-stage detector with feature refinement for rotating object,'' in \emph{Proceedings of the AAAI Conference on Artificial Intelligence}, vol.~35, no.~4, 2021, pp. 3163--3171.

\bibitem{han2021align}
J.~Han, J.~Ding, J.~Li, and G.-S. Xia, ``Align deep features for oriented object detection,'' \emph{IEEE transactions on geoscience and remote sensing}, vol.~60, pp. 1--11, 2021.

\bibitem{yang2019reppoints}
Z.~Yang, S.~Liu, H.~Hu, L.~Wang, and S.~Lin, ``Reppoints: Point set representation for object detection,'' in \emph{Proceedings of the IEEE/CVF international conference on computer vision}, 2019, pp. 9657--9666.

\bibitem{guo2021beyond}
Z.~Guo, C.~Liu, X.~Zhang, J.~Jiao, X.~Ji, and Q.~Ye, ``Beyond bounding-box: Convex-hull feature adaptation for oriented and densely packed object detection,'' in \emph{Proceedings of the IEEE/CVF conference on Computer Vision and Pattern Recognition}, 2021, pp. 8792--8801.

\bibitem{li2022oriented}
W.~Li, Y.~Chen, K.~Hu, and J.~Zhu, ``Oriented reppoints for aerial object detection,'' in \emph{Proceedings of the IEEE/CVF conference on computer vision and pattern recognition}, 2022, pp. 1829--1838.

\bibitem{hou2022shape}
L.~Hou, K.~Lu, J.~Xue, and Y.~Li, ``Shape-adaptive selection and measurement for oriented object detection,'' in \emph{Proceedings of the AAAI Conference on Artificial Intelligence}, vol.~36, no.~1, 2022, pp. 923--932.

\bibitem{tian2019fcos}
Z.~Tian, C.~Shen, H.~Chen, and T.~He, ``Fcos: Fully convolutional one-stage object detection,'' in \emph{Proceedings of the IEEE International Conference on Computer Vision}, 2019, pp. 9627--9636.

\bibitem{yang2020arbitrary}
X.~Yang and J.~Yan, ``Arbitrary-oriented object detection with circular smooth label,'' in \emph{European Conference on Computer Vision}, 2020, pp. 677--694.

\bibitem{yu2024boundary}
Y.~Yu and F.~Da, ``On boundary discontinuity in angle regression based arbitrary oriented object detection,'' \emph{IEEE Transactions on Pattern Analysis and Machine Intelligence}, 2024.

\bibitem{xu2020gliding}
Y.~Xu, M.~Fu, Q.~Wang, Y.~Wang, K.~Chen, G.-S. Xia, and X.~Bai, ``Gliding vertex on the horizontal bounding box for multi-oriented object detection,'' \emph{IEEE transactions on pattern analysis and machine intelligence}, vol.~43, no.~4, pp. 1452--1459, 2020.

\bibitem{han2021redet}
J.~Han, J.~Ding, N.~Xue, and G.-S. Xia, ``Redet: A rotation-equivariant detector for aerial object detection,'' in \emph{Proceedings of the IEEE/CVF conference on computer vision and pattern recognition}, 2021, pp. 2786--2795.

\bibitem{xie2021oriented}
X.~Xie, G.~Cheng, J.~Wang, X.~Yao, and J.~Han, ``Oriented r-cnn for object detection,'' in \emph{Proceedings of the IEEE/CVF International Conference on Computer Vision}, 2021, pp. 3520--3529.

\bibitem{li2023large}
Y.~Li, Q.~Hou, Z.~Zheng, M.-M. Cheng, J.~Yang, and X.~Li, ``Large selective kernel network for remote sensing object detection,'' in \emph{Proceedings of the IEEE/CVF International Conference on Computer Vision}, 2023, pp. 16\,794--16\,805.

\bibitem{cai2024poly}
X.~Cai, Q.~Lai, Y.~Wang, W.~Wang, Z.~Sun, and Y.~Yao, ``Poly kernel inception network for remote sensing detection,'' in \emph{Proceedings of the IEEE/CVF Conference on Computer Vision and Pattern Recognition}, 2024, pp. 27\,706--27\,716.

\bibitem{loshchilov2018decoupled}
I.~Loshchilov and F.~Hutter, ``Decoupled weight decay regularization,'' in \emph{International Conference on Learning Representations}, 2018.

\bibitem{everingham2010pascal}
M.~Everingham, L.~Van~Gool, C.~K. Williams, J.~Winn, and A.~Zisserman, ``The pascal visual object classes (voc) challenge,'' \emph{International Journal of Computer Vision}, vol.~88, pp. 303--338, 2010.

\bibitem{chen2019mmdetection}
K.~Chen, J.~Wang, J.~Pang, Y.~Cao, Y.~Xiong, X.~Li, S.~Sun, W.~Feng, Z.~Liu, J.~Xu \emph{et~al.}, ``Mmdetection: Open mmlab detection toolbox and benchmark,'' \emph{arXiv preprint arXiv:1906.07155}, 2019.

\bibitem{zhou2022mmrotate}
Y.~Zhou, X.~Yang, G.~Zhang, J.~Wang, Y.~Liu, L.~Hou, X.~Jiang, X.~Liu, J.~Yan, C.~Lyu, W.~Zhang, and K.~Chen, ``Mmrotate: A rotated object detection benchmark using pytorch,'' in \emph{Proceedings of the 30th ACM International Conference on Multimedia}, 2022, p. 7331–7334.

\end{thebibliography}

\end{document}